\documentclass[journal]{vgtc}                

\ifpdf
  \pdfoutput=1\relax                   
  \usepackage{graphicx}                
  \DeclareGraphicsExtensions{.pdf,.png,.jpg,.jpeg} 
\else
  \usepackage{graphicx}                
  \DeclareGraphicsExtensions{.eps}     
\fi%

\usepackage{microtype}                 
\PassOptionsToPackage{warn}{textcomp}  
\usepackage{textcomp}                  
\usepackage{mathptmx}                  
\usepackage{times}                     
\usepackage{cite}                      
\usepackage{tabu}                      
\usepackage{booktabs}                  

\onlineid{0}

\vgtccategory{Research}
\vgtcpapertype{please specify}

\usepackage{amsmath}
\usepackage{amssymb}
\usepackage{array}
\usepackage{bm}
\usepackage{fancyhdr}
\usepackage{float}
\usepackage{graphicx}
\usepackage{multirow}
\usepackage{subcaption}

\usepackage{pifont}
\newcommand{\tmark}{\ding{51}}
\newcommand{\xmark}{\ding{55}}

\newcommand{\stufig}[5]                                    
{
	\begin{figure}[#5]
		\begin{center}
			\includegraphics[#1]{#2}
			\caption{#3}
			\label{#4}
		\end{center}
		\vspace{-.5cm}
	\end{figure}
}

\newcommand{\stufigstar}[5]                                
{
	\begin{figure*}[#5]
		\begin{center}
			\includegraphics[#1]{#2}
			\caption{#3}
			\label{#4}
		\end{center}
		\vspace{-.5cm}
	\end{figure*}
}

\newenvironment{stusubfig*}[1]
{
	\begin{figure*}[#1]
		\begin{center}
		}
		{
		\end{center}
	\end{figure*}
}

\DeclareMathOperator*{\argmin}{argmin}

\title{Collaborative Large-Scale Dense 3D Reconstruction \\ with Online Inter-Agent Pose Optimisation}

\author{Stuart Golodetz, Tommaso Cavallari, Nicholas A.\ Lord, \\ Victor A.\ Prisacariu, David W.\ Murray and Philip H.\ S.\ Torr}
\authorfooter{
\item
 This work was done whilst all authors were at the University of Oxford.
\item
 S.\ Golodetz, T.\ Cavallari and N.\ Lord assert joint first authorship.
 They were previously with the University of Oxford, and are now with FiveAI Ltd.
 E-mail: \{stuart,tommaso.cavallari,nick\}@five.ai.
\item
 V.\ Prisacariu, D.\ Murray and P.\ Torr are with the University of Oxford.
 E-mail: \{victor,dwm,phst\}@robots.ox.ac.uk.
}

\shortauthortitle{Biv \MakeLowercase{\textit{et al.}}: Global Illumination for Fun and Profit}

\abstract{\noindent Reconstructing dense, volumetric models of real-world 3D scenes is important for many tasks, but capturing large scenes can take significant time, and the risk of transient changes to the scene goes up as the capture time increases. These are good reasons to want instead to capture several smaller sub-scenes that can be joined to make the whole scene. Achieving this has traditionally been difficult: joining sub-scenes that may never have been viewed from the same angle requires a high-quality camera relocaliser that can cope with novel poses, and tracking drift in each sub-scene can prevent them from being joined to make a consistent overall scene.
Recent advances, however, have significantly improved our ability to capture medium-sized sub-scenes with little to no tracking drift: real-time globally consistent reconstruction systems can close loops and re-integrate the scene surface on the fly, whilst new visual-inertial odometry approaches can significantly reduce tracking drift during live reconstruction.
Moreover, high-quality regression forest-based relocalisers have recently been made more practical by the introduction of a method to allow them to be trained and used online.
In this paper, we leverage these advances to present what to our knowledge is the first system to allow multiple users to collaborate interactively to reconstruct dense, voxel-based models of whole buildings using only consumer-grade hardware, a task that has traditionally been both time-consuming and dependent on the availability of specialised hardware. Using our system, an entire house or lab can be reconstructed in under half an hour and at a far lower cost than was previously possible.
} 


\CCScatlist{ 
 \CCScat{K.6.1}{Management of Computing and Information Systems}%
{Project and People Management}{Life Cycle};
 \CCScat{K.7.m}{The Computing Profession}{Miscellaneous}{Ethics}
}

\teaser{
\begin{subfigure}{.31\linewidth}
	\centering
	\includegraphics[width=\linewidth]{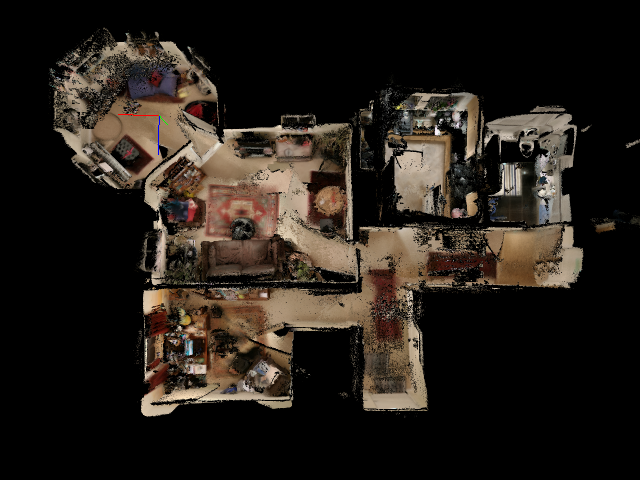}
\end{subfigure}%
\hspace{6mm}%
\begin{subfigure}{.31\linewidth}
	\centering
	\includegraphics[width=\linewidth]{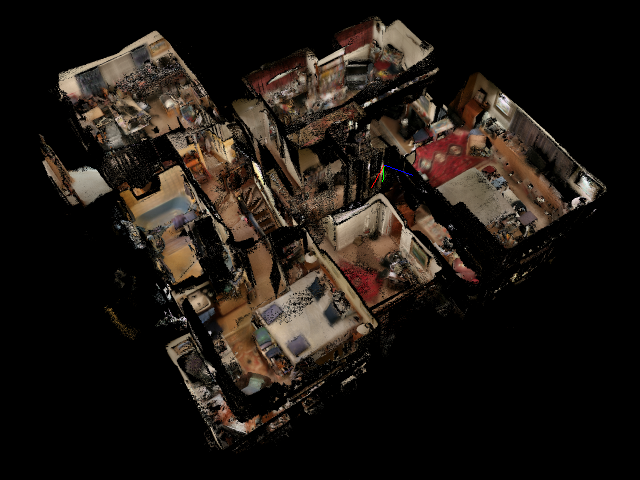}
\end{subfigure}%
\hspace{6mm}%
\begin{subfigure}{.31\linewidth}
	\centering
	\includegraphics[width=\linewidth]{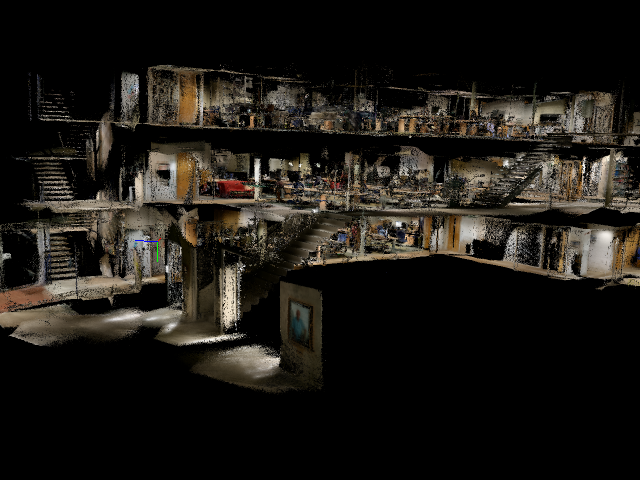}
\end{subfigure}%
\caption{Globally consistent reconstructions produced by our approach, based on the \emph{Flat}, \emph{House} and \emph{Lab} subsets of our dataset.}
\label{fig:teaser}
}

\renewcommand{\manuscriptnotetxt}{\textcopyright{} 2018 IEEE. Personal use of this material is permitted.
	Permission from IEEE must be obtained for all other uses, in any current or future 
	media, including reprinting/republishing this material for advertising or promotional 
	purposes, creating new collective works, for resale or redistribution to servers or 
	lists, or reuse of any copyrighted component of this work in other works. \\
	DOI: \href{https://doi.org/10.1109/TVCG.2018.2868533}{10.1109/TVCG.2018.2868533}}

\vgtcinsertpkg

\begin{document}


\firstsection{Introduction}

\maketitle


\noindent Reconstructing dense, volumetric models of real-world 3D scenes is an important task in computer vision and robotics, with applications in content creation for films and games \cite{Huang2017}, augmented reality \cite{Wasenmueller2016ISMAR}, cultural heritage preservation \cite{Zollhoefer2016} and building information modelling \cite{Murali2017}. Since the seminal KinectFusion work of Newcombe et al.\ \cite{Newcombe2011}, which demonstrated real-time voxel-based reconstruction of a desk-sized scene in real time using only a consumer-grade RGB-D sensor, huge progress has been made in increasing the size of scene we are able to reconstruct \cite{Niessner2013,Prisacariu2014,Steinbruecker2014,Prisacariu2017} and in compensating for tracking drift \cite{Fioraio2015,Whelan2015IJRR,Whelan2015RSS,Kaehler2016,Dai2017}. However, even with the most sophisticated voxel-based approaches \cite{Kaehler2016,Dai2017}, capturing the data needed to reconstruct a large scene (e.g.\ a whole building) can take significant time and planning, and require considerable concentration on the part of the user. Moreover, the risk of transient changes to the scene (e.g.\ people moving around) goes up as the capture time increases, corrupting the model and forcing the user to restart the capture. There are thus good reasons to want to split the capture into several shorter sequences, which can be captured either over multiple sessions or in parallel (by multiple users) and then joined to make the whole scene.

\stufig{width=\linewidth}{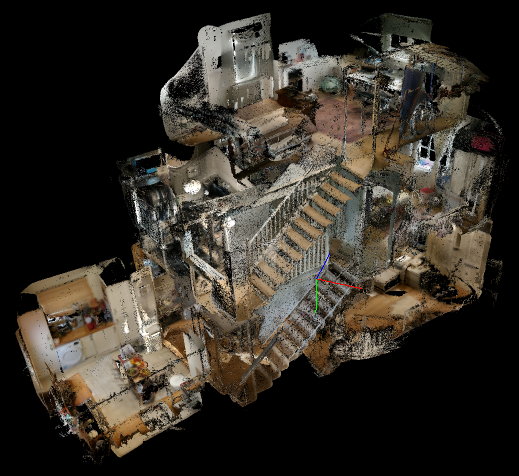}{A globally consistent collaborative reconstruction of a three-storey house, as produced by our approach. The reconstruction involved relocalising the $6$ \emph{Priory} sequences from our dataset.}{fig:example}{!t}

Achieving this has traditionally been difficult: joining the sub-scenes requires the ability to accurately determine the relative transformations between them (a problem that can be expressed as camera relocalisation), even though the areas in which they overlap may never have been viewed from the same angles, and tracking drift in each sub-scene can prevent them from being joined to make a consistent overall scene.
Recent advances, however, have significantly improved our ability to capture consistent, medium-sized sub-scenes, e.g.\ by closing loops and re-integrating the scene surface on the fly \cite{Dai2017}, which yields accurate poses for individual frames once loops have been closed, or by combining visual and inertial cues using an extended Kalman filter \cite{Hesch2014} to achieve accurate camera tracking during live reconstruction.
Moreover, in RGB-D relocalisation, keyframe-based relocalisers such as random ferns \cite{Glocker2015}, which were previously widely used for relocalisation in a single-user context but were unable to relocalise from novel poses, have recently been giving way to high-quality regression-based methods such as SCoRe forests \cite{Shotton2013,Valentin2015RF}, driven by recent work \cite{Cavallari2017} that showed how they could be trained and used online. Unlike keyframe-based methods, such approaches have been shown to be much better suited to relocalisation from novel poses, which is critical when aligning sub-scenes captured from different angles.

In this paper, we leverage these advances to present what to our knowledge is the first system to allow multiple users to collaborate \emph{interactively} to reconstruct dense, voxel-based models of whole buildings. Unlike previous multi-agent mapping approaches \cite{Chebrolu2015,Mohanarajah2015}, our approach is able to reconstruct detailed, globally consistent dense models. Moreover, using our system, an entire house or lab can be captured and reconstructed in under half an hour using only consumer-grade hardware (see \S\ref{subsec:timingexperiments}), providing a low-cost, time-efficient and interactive alternative to existing panorama scanner-based methods for whole-building reconstruction. For example, capturing a building using a Matterport scanner, as done in \cite{Armeni2016}, involves scanning each $2$--$3$m segment of the building separately (taking around $1$--$2$ minutes per segment) and combining the results, a process that can ultimately take many hours. Alternatives such as a Faro 3D laser range finder, as used in \cite{Wijmans2017}, are similarly laborious, with each (large) segment taking up to an hour to reconstruct. Moreover, in both cases, the scanners themselves are quite expensive in comparison to consumer-grade cameras. Our approach, by contrast, is inexpensive in terms of hardware, and allows multiple users to work together interactively to reconstruct the space, significantly reducing the time involved and obviating the need for users to repeatedly move a scanner to reconstruct different segments of a building. A comparison of our approach's properties to those of current methods is shown in Table~\ref{tbl:comparison}. We achieve a $16$-fold increase in scale compared to state-of-the-art low-cost dense solutions such as InfiniTAM \cite{Kaehler2016} and BundleFusion \cite{Dai2017}, making ours the first low-cost, interactive approach that can produce dense reconstructions of this size, and bringing such methods within range of the much more expensive and \emph{offline} alternatives that are currently employed.

We have integrated our approach into the open-source SemanticPaint framework \cite{Golodetz2015SPDEMO,Golodetz2015SPTR}, making it easy for existing SemanticPaint and InfiniTAM \cite{Prisacariu2017} users to benefit from our work, and constructed a new dataset of sub-scenes to demonstrate its effectiveness. We make both our source code and this dataset available online. Figures \ref{fig:teaser} and \ref{fig:example} show reconstructions produced by our approach.

\begin{table}[!t]
	\centering
	\scriptsize
	\begin{tabular}{ccccccc}
		& \cite{Armeni2016} & \cite{Wijmans2017} & \cite{Kaehler2016} & \cite{Dai2017} & Ours \\
		\midrule
		Collaborative & \tmark & \tmark & \xmark & \xmark & \tmark \\
		Interactive & \xmark & \xmark & \tmark & \tmark & \tmark \\
		Low-Cost Hardware & \xmark & \xmark & \tmark & \tmark & \tmark \\
		Largest Reconstruction (m$^2$) & 1900 & 10370 & 50 & 50 & 820 \\
	\end{tabular}
	\caption{A comparison of our approach to state-of-the-art methods for indoor building reconstruction. `Largest Reconstruction' refers to the size of the largest reconstruction demonstrated in the relevant papers. For our method, $820$m$^2$ refers to a collaborative reconstruction of the \emph{Lab} subset of our dataset, and was obtained by cross-referencing our reconstruction with an official floorplan.}
	\label{tbl:comparison}
	\vspace{-\baselineskip}
\end{table}

\section{Related Work}

\stufigstar{width=\linewidth}{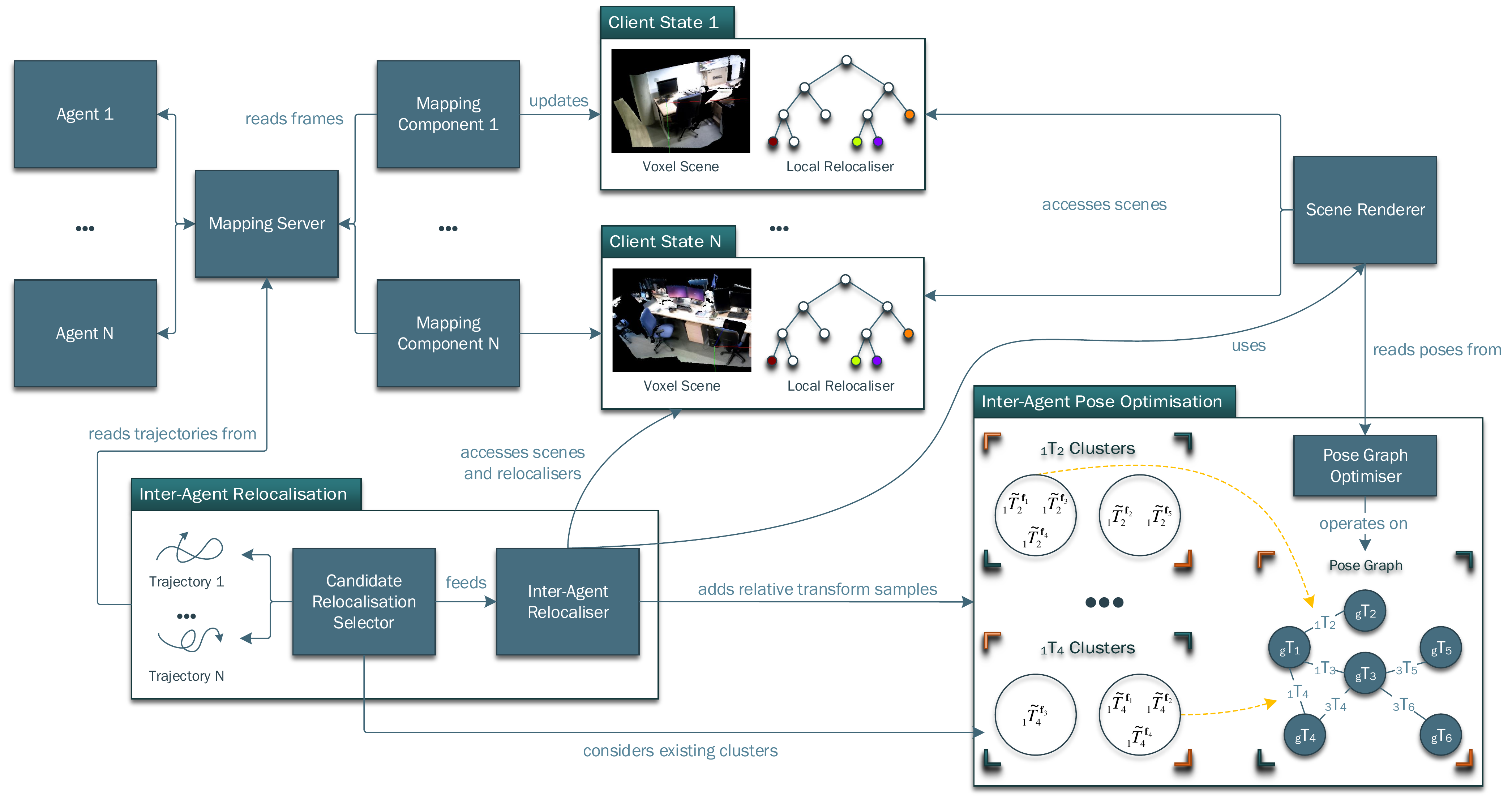}{The architecture of our system. Individual agents track their poses and feed posed \mbox{RGB-D} frames to the mapping server. A separate mapping component is instantiated for each agent, which reconstructs a voxel scene and trains a local relocaliser. Separately, the candidate relocalisation selector repeatedly selects a pose from one of the agents' trajectories for relocalisation against another scene. The inter-agent relocaliser uses the scene renderer to render synthetic RGB and depth images of the corresponding scene from the selected pose, and passes them to the local relocaliser of the target scene. If a relocalisation succeeds (and is verified, see \S\ref{subsubsec:interagentrelocalisation}), a sample of the relative transform between the two scenes is recorded. The relative transform samples for each scene pair are clustered for robustness (see \S\ref{subsubsec:interagentposeoptimisation}). Whenever the cluster to which a sample is added is sufficiently large, we construct a pose graph by blending the relative poses in the largest clusters, and trigger a pose graph optimisation. The optimised poses are then used for rendering the overall scene.}{fig:pipeline}{!t}

Although previous work on multi-agent mapping has not focused on dense reconstruction, multi-agent mapping itself has a rich research history in computer vision and robotics, and several good surveys exist \cite{Rone2013,Saeedi2016}.
Existing approaches occupy two main categories:

(i) \emph{Decentralised} approaches eschew the use of a central server and instead produce a local map of the scene on each agent, often transmitting these local maps between agents when they meet to share knowledge of parts of the scene that individual agents have not yet visited. For example, Cunningham et al.\ \cite{Cunningham2010} proposed an approach called DDF-SAM, in which each robot produces a landmark-based map and shares compressed, timestamped versions of it with neighbouring robots. This was extended by \cite{Cunningham2012}, which registered landmark-based maps together using an approach based on Delaunay triangulation and RANSAC. The same lab later proposed DDF-SAM 2.0 \cite{Cunningham2013}, which avoids repeated and expensive recreation of the combined neighbourhood map through the use of `anti-factors'.
Cieslewski et al.\ \cite{Cieslewski2015} presented a sophisticated decentralised collaborative mapping back-end based on distributed version control. Choudhary et al.\ \cite{Choudhary2017} performed decentralised mapping using object-based maps, decreasing the bandwidth required for map sharing, but depending on the existence of pre-trained objects in the scene. Most recently, Cieslewski et al.\ \cite{Cieslewski2017} aimed to minimise the bandwidth each robot uses for inter-agent relocalisation by first establishing which other robots have relevant information, and then only communicating with the best robot found. Decentralised approaches have numerous applications, including search and rescue \cite{Michael2012}, agricultural robotics \cite{Cheein2013}, planetary exploration \cite{Bajpai2016} and underwater mapping \cite{Paull2015,Silveira2015}, but because of the limited computing power that tends to be available on mobile agents, most approaches target robustness to unreliable network connections and mechanical failures, rather than reconstructing detailed scene geometry, limiting their use for tasks like content creation, building information modelling or cultural heritage preservation.

(ii) \emph{Centralised} approaches, by contrast, can take advantage of the computing power provided by one or more central servers and their ability to communicate with all agents at once to produce detailed, globally consistent maps. For example, Chebrolu et al.\ \cite{Chebrolu2015} described a semi-dense approach based on LSD-SLAM \cite{Engel2014} in which monocular clients produced pose-graph maps of keyframes that were sent to a central server for optimisation.
Riazuelo et al.\ \cite{Riazuelo2014} described a cloud-based distributed visual SLAM system inspired by PTAM \cite{Klein2007}. Their clients performed tracking only, with the more expensive mapping steps being performed in the cloud.
Mohanarajah et al.\ \cite{Mohanarajah2015} described another cloud-based approach based on the authors' Rapyuta robotics platform. The client robots estimated their local poses by running dense, keyframe-based visual odometry on RGB-D images from a PrimeSense camera. The keyframes were then centrally optimised using g$^2$o \cite{Kummerle2011}.
Poiesi et al.\ \cite{Poiesi2017} described yet another cloud-based approach that performs Structure from Motion (SfM) and local bundle adjustments on monocular videos from smartphones to reconstruct a consistent point cloud map for each client, and more costly periodic full bundle adjustments to align the maps of different clients.
Forster et al.\ \cite{Forster2013} demonstrated centralised, keyframe-based collaborative mapping from micro aerial vehicles (MAVs), each equipped with a monocular camera and an IMU.
More recently, Schmuck and Chli \cite{Schmuck2017} have shown how to incorporate server-to-client feedback into a multi-UAV collaborative approach, allowing agents to share information.

A few approaches do not fit cleanly into either category. Reid et al.\ \cite{Reid2013} described a distributed approach in which multiple autonomous ground robots were controlled from a centralised ground control station (GCS), but were able to operate independently for a while if the connection to the GCS failed. McDonald et al.\ \cite{McDonald2013} described a stereo approach in which a single agent reconstructed a scene over multiple sessions, effectively collaborating with itself over time. Chen et al.\ \cite{Chen2014} described an approach that is initially peer-to-peer, with each robot building a pose graph independently, storing its sensor data, and exchanging information with other robots on rendezvous, but later client-server, with the robots transferring their pose graphs and sensor data to a central server for pose graph optimisation and the building of a coloured point cloud map. Fankhauser et al.\ \cite{Fankhauser2016} described an approach that allowed an Asctec Firefly hexacopter and a quadrupedal ground robot to work together to help the ground robot to navigate safely. Whilst no fixed-position central server was used, the ground robot had significant computing power and effectively played the role of a server, bundle adjusting data from the hexacopter to maintain a globally consistent map.

\section{Our Approach}

Since we target scenarios such as indoor building reconstruction, where hardware failure is a minor concern, we adopt a centralised approach with multiple clients and a powerful central server (e.g.\ a laptop or desktop with one or more high-end GPUs): see Figure~\ref{fig:pipeline}. This gives us the computational power to relocalise different clients' sub-scenes against each other fast enough to support interactive reconstruction, and reduces the network bandwidth we require (since there is then no need for direct peer-to-peer communication). Moreover, since interactive reconstruction is a key target of our approach, we stipulate that each client must estimate accurate local poses for a sequence of RGB-D frames (see \S\ref{subsec:localposeestimation}) and transmit both the frames and the poses to the central server (see \S\ref{subsec:frametransmission}). We do this for two reasons: (i) it is helpful to allow users to see a good approximation of the final reconstruction interactively to help guide the mapping process, and (ii) insisting that the local submaps are consistent during live reconstruction is essential if we are to generate a consistent overall map on the fly. The ability to provide a consistent live map has important implications for applications such as robotics, since a robot that cannot trust its map must waste computation on modelling this uncertainty when making decisions.

For interactive reconstruction, our insistence on accurate local poses implies the client-side use of a high-quality camera tracker (e.g.\ one based on visual-inertial odometry \cite{Hesch2014}) that can provide such poses on the fly. However, our approach is general enough that we can also support an \emph{offline} (batch) reconstruction mode. For this, accurate live local poses are not strictly necessary, allowing more flexible schemes to be used. In particular, one option (see \S\ref{subsec:adaptabilityexperiments}) is to reconstruct a submap for each client using BundleFusion \cite{Dai2017} and then send the bundle-adjusted poses to the server. Alternatively, the clients could transmit inaccurate local poses and the bundle adjustment could be performed on the server, although that has some downsides (see \S\ref{subsec:localposeestimation}) and would involve a different architecture.

In our architecture, regardless of how the clients are implemented, the server constructs a sub-scene and trains a relocaliser for each client based on the frames and poses it receives (however they are produced), and relocalises the sub-scenes from different clients against each other to establish a consistent global map (see \S\ref{subsec:globalmapping}).
For a batch reconstruction, all of the sub-scenes are sent across to the server (or simply read directly from disk) before any relocalisation is performed.
For an interactive reconstruction, the server will start relocalising clients against each other immediately, and new clients can join on the fly to contribute to the map.
During the reconstruction process itself, the sub-scenes are kept separate to make it possible to continually optimise the relative poses between them (see \S\ref{subsubsec:interagentposeoptimisation}); if desired, they can later be fused into a single consistent map at the end of the reconstruction process, as described in \S\ref{subsec:singleglobalreconstruction}.

In practice, it can be helpful for collaborating clients to see which parts of the space have already been reconstructed by other users. For this reason, our approach allows each client to ask the server to render the global scene from a specified camera pose (see \S\ref{subsec:globalmapfeedback}). This server-to-client feedback is particularly helpful in situations in which users are out of each other's visual range, since it can help them (i) plan how they will meet to join their sub-scenes together, and (ii) decide which parts of the space still need to be mapped.

\subsection{Local Pose Estimation (Client-Side)}
\label{subsec:localposeestimation}

As mentioned above, each client in our system must estimate accurate local poses for a sequence of RGB-D frames and then transmit both the frames and the poses to the central server. This is non-trivial, since traditional pose estimation approaches, particularly those based purely on visual tracking, have tended to be subject to significant tracking drift, particularly at larger scales. For collaborative mapping, a common solution has been for clients to simply transmit inaccurate local poses and rely on the server to perform global optimisation (e.g.\ pose graph optimisation between keyframes \cite{Reid2013,Chen2014,Mohanarajah2015} or bundle adjustment \cite{Chebrolu2015,Fleck2015,Schmuck2017}) to achieve a globally consistent map (the optimised poses can then be sent back to the clients if desired \cite{Chebrolu2015,Mohanarajah2015,Schmuck2017}). However, (i) such global optimisations do not scale well, limiting the overall size of map that can be constructed, and (ii) the local poses on the clients are only corrected as global optimisations on the server finish, meaning that for much of the time they cannot be fully trusted. For these reasons, we choose in our approach to place the burden of accurate local pose estimation on the clients, since (i) that is a more natural fit for the interactive scenario we target, and (ii) it frees the server to focus on reconstructing a dense global map in real time. Two side benefits of this approach are that it significantly simplifies the design of the server (see \S\ref{subsec:globalmapping}) and reduces server-side memory consumption, making it possible to support more agents and larger global maps.

In practice, our approach is agnostic to how a client that can satisfy this requirement is implemented, and many popular solutions (generally based on some form of visual-inertial odometry) exist \cite{Hesch2014,Bloesch2015,ARCore,ARKit} (for concrete examples, see \S\ref{sec:experiments}).
Moreover, our approach naturally supports collaborations between different types of client, making it possible for them to cooperate on the same reconstruction.

\subsection{RGB-D Frame Transmission (Client $\rightarrow$ Server)}
\label{subsec:frametransmission}

We maintain a TCP connection between each client and the server to guarantee in-order delivery of RGB-D frames. To minimise the network bandwidth, we only transmit frames that were tracked successfully, and we compress the depth images in PNG format (lossless) and the RGB images in JPG format (lossy). Moreover, to maintain a smooth, interactive user experience on the client, we transmit messages containing the frames (and their accompanying poses) to the server on a separate thread, which iteratively reads a frame message from a \emph{pooled queue} of reusable messages, compresses it and sends it to the server.\footnote{Further details about the design of the pooled queue data structure can be found in \S\ref{subsec:pooledqueue}.}
The main thread writes uncompressed frame messages into this queue based on the current input.
We discard messages that would overflow the queue (e.g.\ when the network cannot keep up with the client) to maintain interactivity and bound client memory usage; the way in which this interacts with our compression strategy is evaluated in \S\ref{subsubsec:effectiveness-framecompression}.

At the server end, each client has a handler (running on a separate thread) that maintains a pooled queue of uncompressed frame messages. When a compressed frame message arrives, it is immediately uncompressed and pushed onto the queue (as on the client, we discard messages that would overflow the queue). On the main thread, we run a mapping component for each client (see \S\ref{subsec:globalmapping}), which reads \mbox{RGB-D} frames and their accompanying poses from its client's queue as necessary and creates a local sub-scene.

\subsection{Global Mapping (Server-Side)}
\label{subsec:globalmapping}

The server has two jobs: (i) constructing a sub-scene and training a relocaliser for each client, and (ii) determining the relative transforms between these sub-scenes to establish a global coordinate system. To achieve the first, a separate mapping component for each client (a) runs the open-source InfiniTAM reconstruction engine \cite{Prisacariu2014,Prisacariu2017} on incoming (posed) \mbox{RGB-D} frames to construct a voxel-based map, and (b) trains a regression forest-based relocaliser for the sub-scene online as per Cavallari et al.~\cite{Cavallari2017}. To achieve the second, the server attempts to relocalise synthetic images of one agent's sub-scene using another agent's relocaliser to find estimates of the relative transform between the sub-scenes (see \S\ref{subsubsec:interagentrelocalisation}). These samples are clustered in transformation space to help suppress outliers, and
a pose graph is constructed and optimised in the background to further refine the relative transforms between the sub-scenes (see \S\ref{subsubsec:interagentposeoptimisation}). This optimisation is inspired by the single-agent sub-mapping approach of K{\"a}hler et al.\ \cite{Kaehler2016}, which showed how to build globally consistent voxel-based models by dividing a scene into small sub-scenes and optimising the relative poses between them. However, we construct a pose graph where each agent is represented by a single node, and the edges denote the relative transforms between different agents' sub-scenes. This differs from \cite{Kaehler2016}, where all sub-scenes came from one agent and no relocalisation was needed to establish the transforms between them.

\subsubsection{Inter-Agent Relocalisation}
\label{subsubsec:interagentrelocalisation}

\stufigstar{width=.9\linewidth}{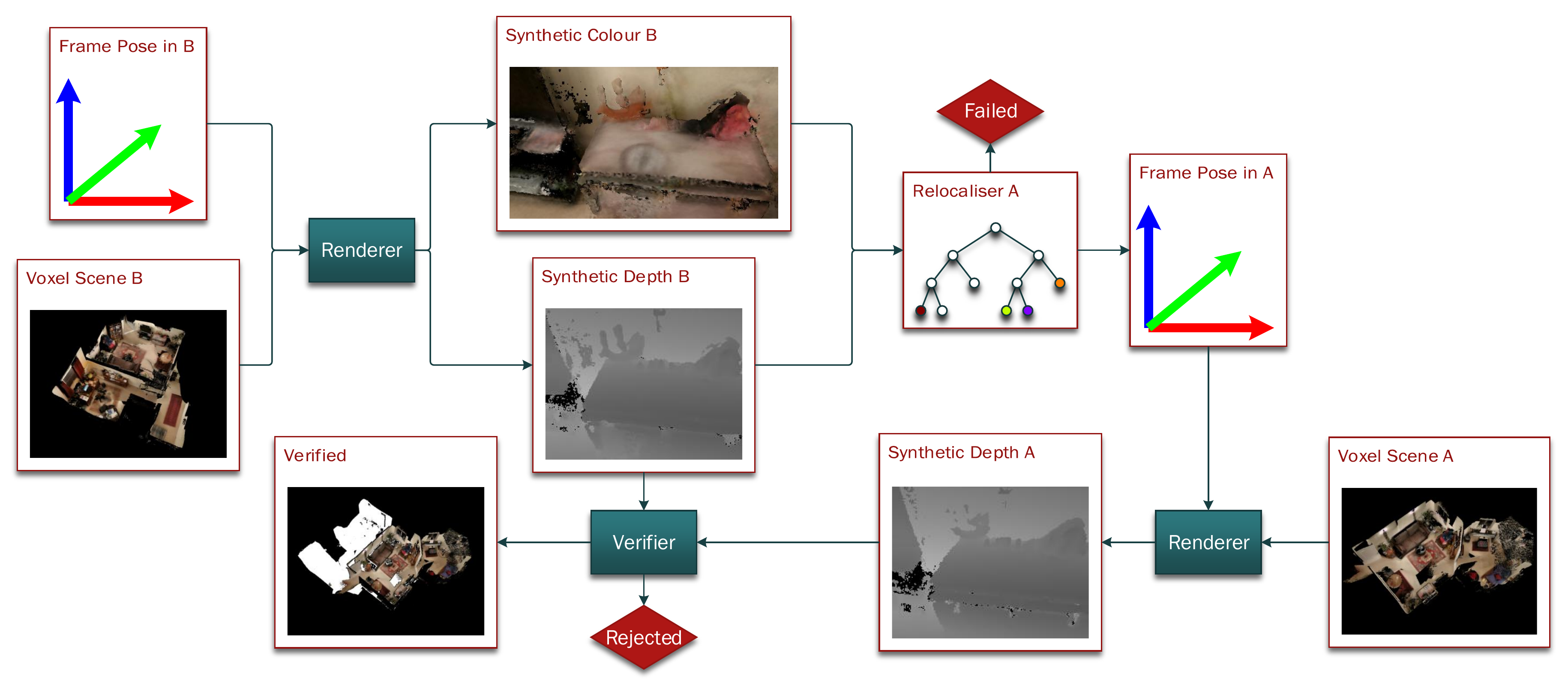}{To relocalise the scene of an agent $b$ against that of another agent $a$, we first choose an arbitrary frame from $b$'s trajectory and render synthetic RGB and depth raycasts of $b$'s scene from the frame's pose. Then, we try to relocalise them using $a$'s relocaliser, which either fails, or produces an estimated pose for the frame in $a$'s coordinate system. If a pose is proposed, we verify it by rendering a synthetic depth raycast of $a$'s scene from the proposed pose and comparing it to the synthetic depth raycast of $b$'s scene. We accept the pose if and only if the two depth raycasts are sufficiently similar (see \S\ref{subsubsec:interagentrelocalisation}).}{fig:relocalisation}{!t}

To maintain a smooth, interactive experience on the server, we attempt relocalisations between the sub-scenes of different clients on a separate thread and (if available) a separate GPU. The scheduling of relocalisation attempts depends on the server mode. In \emph{batch} mode, relocalisations are attempted only once all of the client sub-scenes have been fully created, at which point they are attempted as quickly as possible (i.e.\ a new relocalisation attempt is scheduled as soon as the previous attempt finishes). In \emph{interactive} mode, we relocalise whilst the client sub-scenes are still being reconstructed, but at most once every $50$ frames: we do this to space out relocalisation attempts and allow sufficient time in between attempts for the relocalisers to be trained.

To schedule an attempt, we first randomly generate a list of $10$ candidate relocalisations. Each candidate $\mathbf{k} = (a,\mathbf{f})$, where $\mathbf{f} = (b,i)$, denotes an attempt to relocalise frame $i$ of scene $b$ in scene $a$'s coordinate system. To balance different pairs of scenes, which may have been reconstructed from varying numbers of frames, we first uniformly sample an $(a,b)$ scene pair from the set of all scene pairs, and then uniformly sample a frame index $i$ from scene $b$. Each generated candidate $\mathbf{k} = (a,(b,i))$ is then scored via
\begin{equation}
\footnotesize
\Phi(\mathbf{k}) = \phi_\mathit{new}(\mathbf{k}) - \phi_\mathit{conf}(\mathbf{k}) - \phi_\mathit{homog}(\mathbf{k}).
\end{equation}
In this, $\phi_\mathit{new}$ aims to give a boost to candidates that might connect new nodes to the pose graph, defined as $1$ if one of the sub-scenes $a$ and $b$ already has an optimised global pose and the other does not, and $0$ otherwise. $\phi_\mathit{conf}$ penalises candidates that can only add a relative transform sample between a pair of sub-scenes that are already confidently relocalised with respect to each other:
\begin{equation}
\footnotesize
\phi_\mathit{conf}(a,(b,i)) = \max{}\left(0, \max_{c \in \mathit{clusters}(a,b)} (|c| - N)\right).
\end{equation}
Here, $N = 2$ is an empirically-chosen threshold on the number of relocalisations needed before we become confident that the relative transform between a pair of sub-scenes is correct. $\phi_\mathit{homog}$ penalises candidates whose frame $i$ has a local pose in scene $b$ that is too close to (within $5\mbox{cm}$ / $5^\circ$ of) one that has already been tried against scene $a$; we use a penalty of $5$ for poses that are too close, and $0$ otherwise.

Having scored the candidates, we schedule a relocalisation attempt for a candidate $\mathbf{k} = (a,\mathbf{f})$ with maximum score, where $\mathbf{f} = (b,i)$. This proceeds as shown in Figure~\ref{fig:relocalisation}. Let ${}_\mathbf{f}\tilde{T}_s$ denote the pose of frame $\mathbf{f}$ in the coordinate system of a sub-scene $s$. First, we render synthetic RGB and depth raycasts $\tilde{C}_b^\mathbf{f}$ and $\tilde{D}_b^\mathbf{f}$ of sub-scene $b$ from the known pose ${}_\mathbf{f}\tilde{T}_b$. Then, we try to relocalise them using sub-scene $a$'s relocaliser to obtain an estimated pose ${}_\mathbf{f}\tilde{T}_a$ for $\mathbf{f}$ in $a$'s coordinate system. The effectiveness of relocalising from synthetic images is evaluated in \S\ref{subsubsec:effectiveness-synthetic}.

To verify the estimated pose, we render a synthetic depth raycast $\tilde{D}_a^\mathbf{f}$ of sub-scene $a$ from ${}_\mathbf{f}\tilde{T}_a$, and compare it to $\tilde{D}_b^\mathbf{f}$. To perform the comparison, we let $\Omega$ be the common domain of $\tilde{D}_a^\mathbf{f}$ and $\tilde{D}_b^\mathbf{f}$ (i.e.\ the set of possible pixel coordinates in the two images), $\Omega_s$ be the subset of $\Omega$ for which $\tilde{D}_s^\mathbf{f}(\mathbf{x})$ is valid (where $s$ is either $a$ or $b$), i.e.
\begin{equation}
\footnotesize
\Omega_s = \{\mathbf{x} \in \Omega: \tilde{D}_s^\mathbf{f}(\mathbf{x}) > 0\},
\end{equation}
and $\Omega_{a,b} = \Omega_a \cap \Omega_b$. Then, provided $|\Omega_a| / |\Omega|$ is greater than a threshold (empirically set to $0.5$), i.e.\ provided the estimated pose ${}_\mathbf{f}\tilde{T}_a$ in $a$'s coordinate system is plausible, in the sense that it points sufficiently towards the reconstructed sub-scene, we compute a mean (masked) absolute depth difference between $\tilde{D}_a^\mathbf{f}$ and $\tilde{D}_b^\mathbf{f}$ via
\begin{equation}
\footnotesize
\mu = \left(\sum_{\mathbf{x} \in \Omega_{a,b}} \left|\tilde{D}_a^\mathbf{f}(\mathbf{x}) - \tilde{D}_b^\mathbf{f}(\mathbf{x})\right|\right) / |\Omega_{a,b}|,
\end{equation}
and add a relative transform sample ${}_a\tilde{T}_b^\mathbf{f} = ({}_\mathbf{f}\tilde{T}_a)^{-1} {}_\mathbf{f}\tilde{T}_b$ between $a$ and $b$ if and only if $\mu$ is sufficiently small (empirically, we found $5$cm to be a good threshold). The effectiveness of this step is evaluated in \S\ref{subsubsec:effectiveness-verification}.

\subsubsection{Inter-Agent Pose Optimisation}
\label{subsubsec:interagentposeoptimisation}

We incrementally cluster the relative transform samples $\{{}_a\tilde{T}_b^\mathbf{f}\}$ we add for each pair of sub-scenes $(a,b)$ prior to performing pose graph optimisation to suppress the effect of outliers on the final result. After adding each sample, we look to see if the cluster to which the sample has just been added now has $\ge N$ samples (i.e.\ the sample has contributed to a confident relative transform): if so, it is worthwhile to run pose graph optimisation, since the pose graph we construct may have changed since the last run.

To construct the pose graph $G$, we first construct an intermediate graph $G'$ that has a node for every sub-scene and an edge connecting every pair of sub-scenes between which a confident relative transform exists (i.e.\ every pair whose largest cluster has size $\ge N$).
For each edge $(a,b)$ in $G'$, we blend the relative transform samples $\{{}_a\tilde{T}_b^\mathbf{f}\}$ in the largest cluster using dual quaternion blending \cite{Kavan2006} to form an overall estimate ${}_a\tilde{T}_b$ of the relative transform from $b$ to $a$.
Next, we compute the connected components of $G'$, such that two sub-scenes end up in the same connected component if and only if there is a chain of confident relative transforms between them.
Finally, we set the pose graph $G$ to be the connected component of $G'$ containing the sub-scene corresponding to the first agent.
Note that, by design, $G$ is quite small, containing only one edge for each sub-scene pair: this is helpful in our context, since it allows the optimisation to be run repeatedly in the background to optimise the overall map.

The goal of optimising $G$ is to find an optimised global pose ${}_g\hat{T}_a$ for each sub-scene $a$ for which $G$ has a node. To perform the optimisation, we use the approach of K{\"a}hler et al.\ \cite{Kaehler2016}, as implemented in InfiniTAM \cite{Prisacariu2014,Prisacariu2017}. This uses Levenberg-Marquardt to minimise the error function
\begin{equation}
\footnotesize
\epsilon(G) = \sum_{(a,b) \in \mathit{edges}(G)} \left\| \mathbf{v}({}_g\hat{T}_b^{-1} \; {}_g\hat{T}_a \; {}_a\tilde{T}_b) \right\|_2,
\end{equation}
where $\mathbf{v}(T) = (\hat{\mathbf{q}}(T), \mathbf{t}(T))^\top$ denotes the concatenation of $\hat{\mathbf{q}}(T)$, the three imaginary components of a quaternion representing the rotational part of $T$, and $\mathbf{t}(T)$, the translational part of $T$. Implicitly, this optimisation is trying to achieve ${}_g\hat{T}_a^{-1} \; {}_g\hat{T}_b = {}_a\tilde{T}_b$ for every $a$ and $b$, i.e.\ to ensure that the optimised global poses of all the scenes are consistent with the estimated relative poses between them.

\subsection{Global Map Feedback (Server $\rightarrow$ Client)}
\label{subsec:globalmapfeedback}

To allow users to collaborate more effectively to reconstruct a scene (particularly in situations in which they can be out of each other's visual range for at least some of the time), we allow each client to ask the server to render the global scene from a single specified camera pose: the rendered images can then be shown to the users to help them plan how to join their sub-scenes together and decide which parts of the space still need to be mapped. As when transmitting frames in the client-to-server direction, we compress the rendered images before sending them to the clients via TCP.

To achieve acceptable performance for multiple clients, we render low-resolution images and service only the most recent request from each client. Specifically, the server stores a pose from which to render for each client, and has a background thread that repeatedly renders an image for one client at a time (from its desired rendering pose), in a round-robin fashion to ensure fairness between the different clients. Each client can update its desired rendering pose at any point by sending a pose update message to the server (in practice, we send such messages once per frame). The reason for adopting this approach is to reduce network bandwidth and maintain interactivity on both the server and the clients, at the cost of some client-side rendering latency (specifically, the global scene image we show on each client will generally be slightly out of date in comparison to the most recent version of the global scene on the server). For live collaborative reconstruction, we found that this latency was not a major problem in practice: clients simply need to have an idea of what the global scene looks like so that they can plan what actions to take next, and those actions do not generally vary significantly in response to minor changes in the global scene.

\section{Experiments}
\label{sec:experiments}

We perform both quantitative and qualitative experiments to evaluate our approach. As mentioned in \S\ref{subsec:localposeestimation}, our approach is able to work with any RGB-D client that can provide sufficiently accurate local poses to the server. To demonstrate this, we implemented two different types of client: (i) a straightforward Android client that captures and streams the \mbox{RGB-D} images and poses provided by the visual-inertial odometry solution in an Asus ZenFone AR augmented reality smartphone, and (ii) a PC-based client integrated into SemanticPaint \cite{Golodetz2015SPTR}. By using both visual and inertial cues, the ZenFone client is able to provide high-quality poses, but camera limitations mean that the depth provided is fairly low resolution (only $224 \times 172$). By contrast, the PC-based client can be used with conventional RGB-D cameras such as a Kinect, PrimeSense or Orbbec Astra, which can provide higher-resolution depth (e.g.\ $640 \times 480$), but the visual-only tracking provided in the InfiniTAM framework \cite{Prisacariu2017} on which SemanticPaint is based can drift far more than visual-inertial odometry, particularly at larger scales.

One option for mitigating this is to capture data using a real-time globally consistent reconstruction system such as BundleFusion \cite{Dai2017}, which can close loops and re-integrate the scene surface on the fly, and then use our PC-based client to feed frames with bundle-adjusted poses to the server. However, BundleFusion can only provide accurate poses once loops have been closed (limiting its use to collaboration in batch mode), and its visual-only tracker provides poses that, even after loop closure resolution, are less accurate than those provided by visual-inertial odometry. Moreover, such trackers can sometimes fail before loops have been closed, e.g.\ due to untextured regions or a lack of interesting geometry, in which case reconstruction will fail. Since reliable tracking and accurate local poses are important in our large-scale context, we used the Android client for most of our experiments and to create our dataset. (On a practical level, smartphones are also more convenient for collaborative reconstruction than heavy laptops.) Nevertheless, to demonstrate our approach's adaptability to other client implementations, we show a collaborative reconstruction using BundleFusion in \S\ref{subsec:adaptabilityexperiments}.

In overview, we conduct the following experiments. In \S\ref{subsec:qualityexperiments}, we evaluate the quality of a reconstruction we obtain with our collaborative approach.
In \S\ref{subsec:effectivenessexperiments}, we evaluate the effectiveness of various individual components of our system.
In \S\ref{subsec:scalabilityexperiments}, we demonstrate the scalability of our approach in terms of both time and memory usage.
In \S\ref{subsec:timingexperiments}, we time how long our approach takes to produce consistent reconstructions for four different subsets of our dataset. We show in particular that using our system, it is possible to collaboratively reconstruct an entire three-storey research lab in under half an hour.
Finally, in \S\ref{subsec:adaptabilityexperiments}, we show the adaptability of our approach to different client implementations, as previously mentioned.
See \S\ref{sec:dataset} for further details about our dataset.

\subsection{Reconstruction Quality}
\label{subsec:qualityexperiments}

\stufig{width=\linewidth}{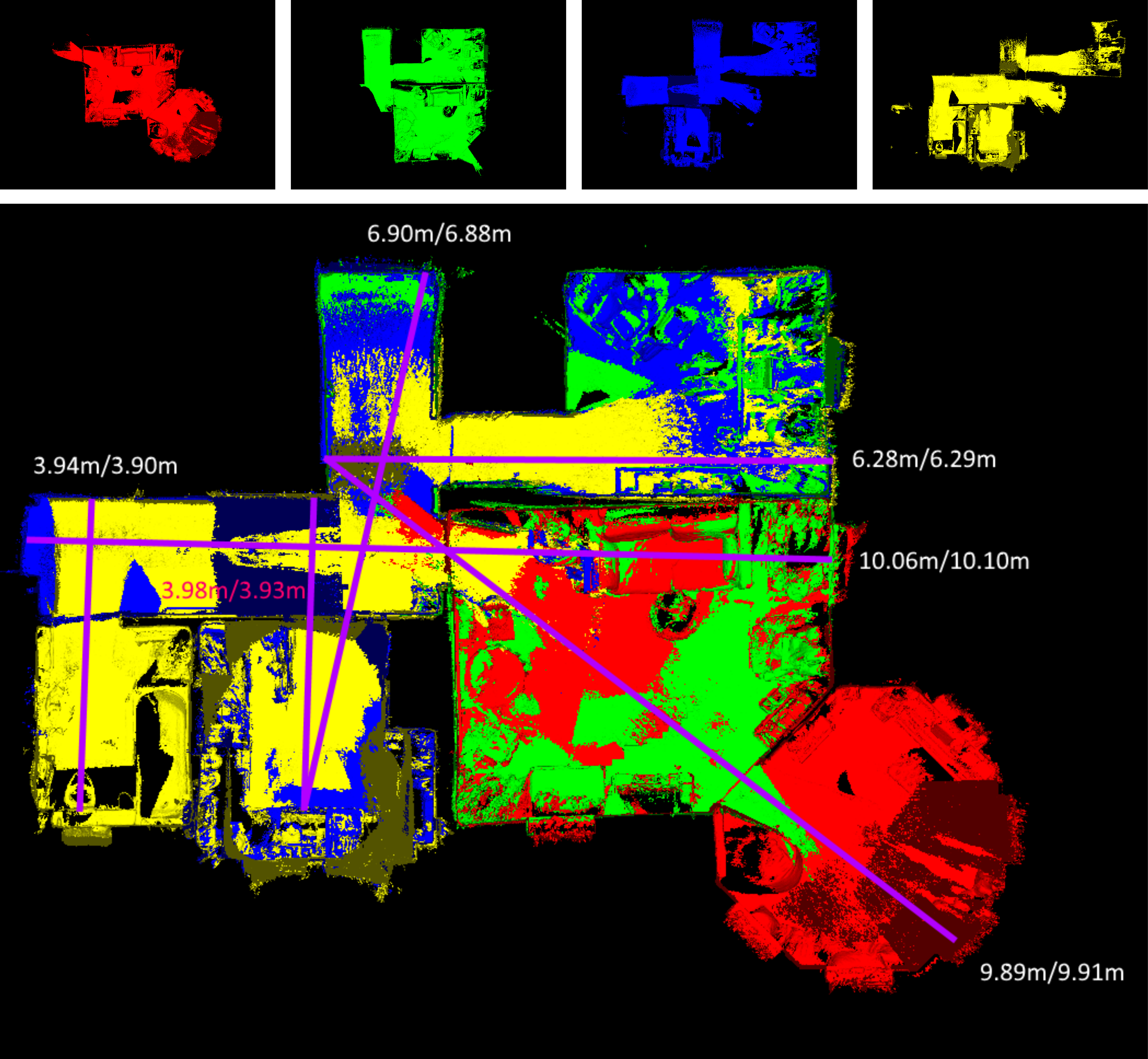}{An example of the high-quality reconstructions we can achieve using our collaborative approach. We joined the $4$ \emph{Flat} sequences from our dataset to make a combined map of a two-bedroom flat. The purple lines show the measurements we performed both using a laser range finder (in the real world) and on a combined mesh of our model (in MeshLab \cite{Cignoni2008}) to validate our approach (the ordering of the numbers is range finder then mesh).}{fig:reconstructionquality}{!t}

\begin{stusubfig*}{!t}
	\begin{subfigure}{.26\linewidth}
		\centering
		\includegraphics[width=\linewidth]{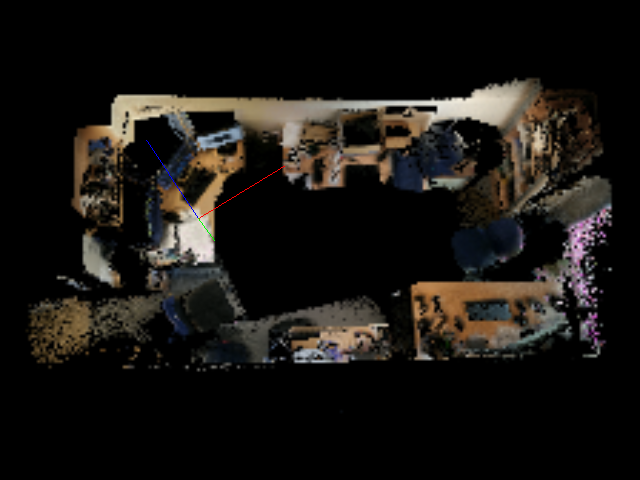}
		\caption{Reference Model}
	\end{subfigure}%
	\hspace{4mm}%
	\begin{subfigure}{.32\linewidth}
		\centering
		\includegraphics[width=\linewidth]{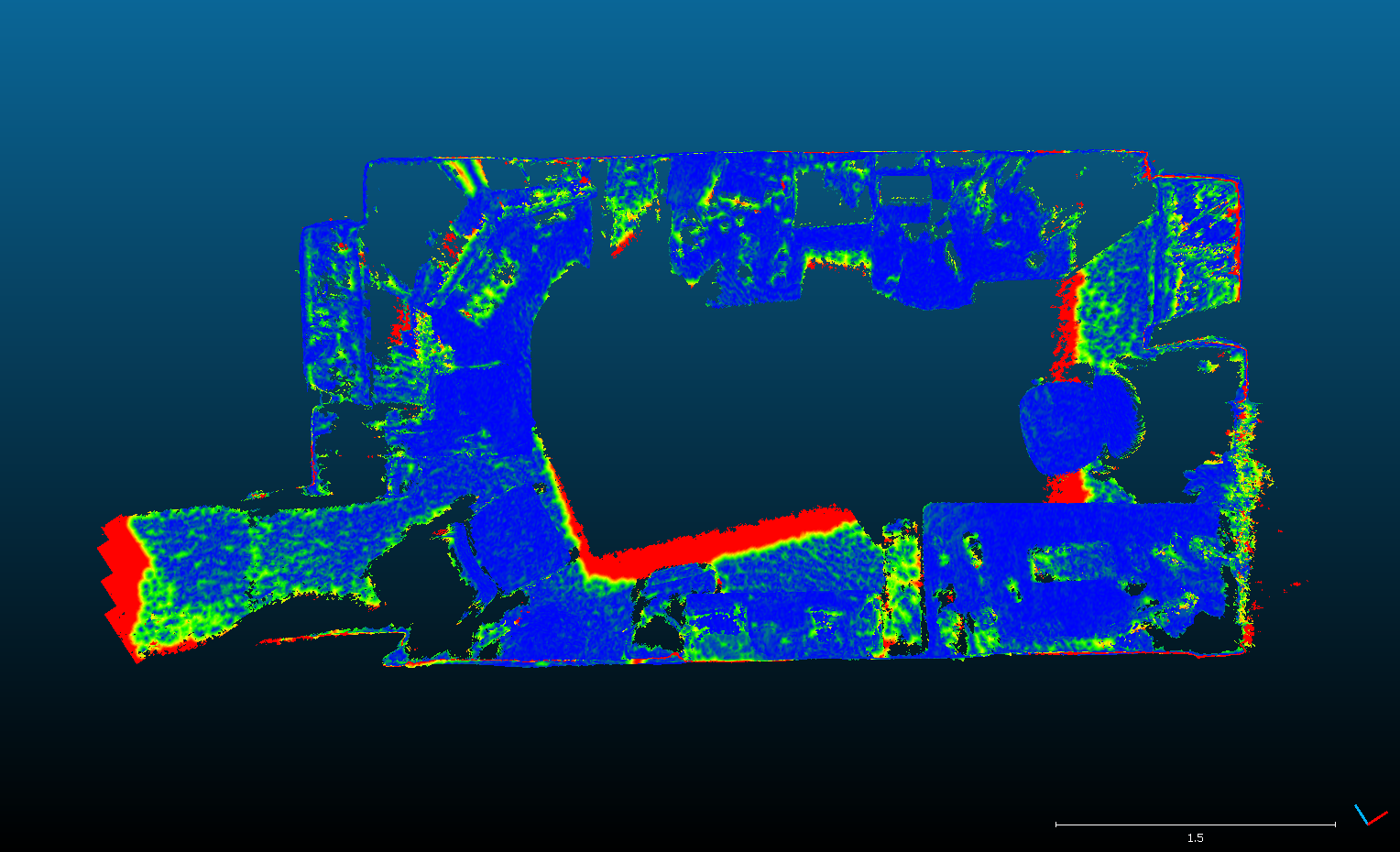}
		\caption{Reference vs.\ Uncompressed}
	\end{subfigure}%
	\hspace{4mm}%
	\begin{subfigure}{.32\linewidth}
		\centering
		\includegraphics[width=\linewidth]{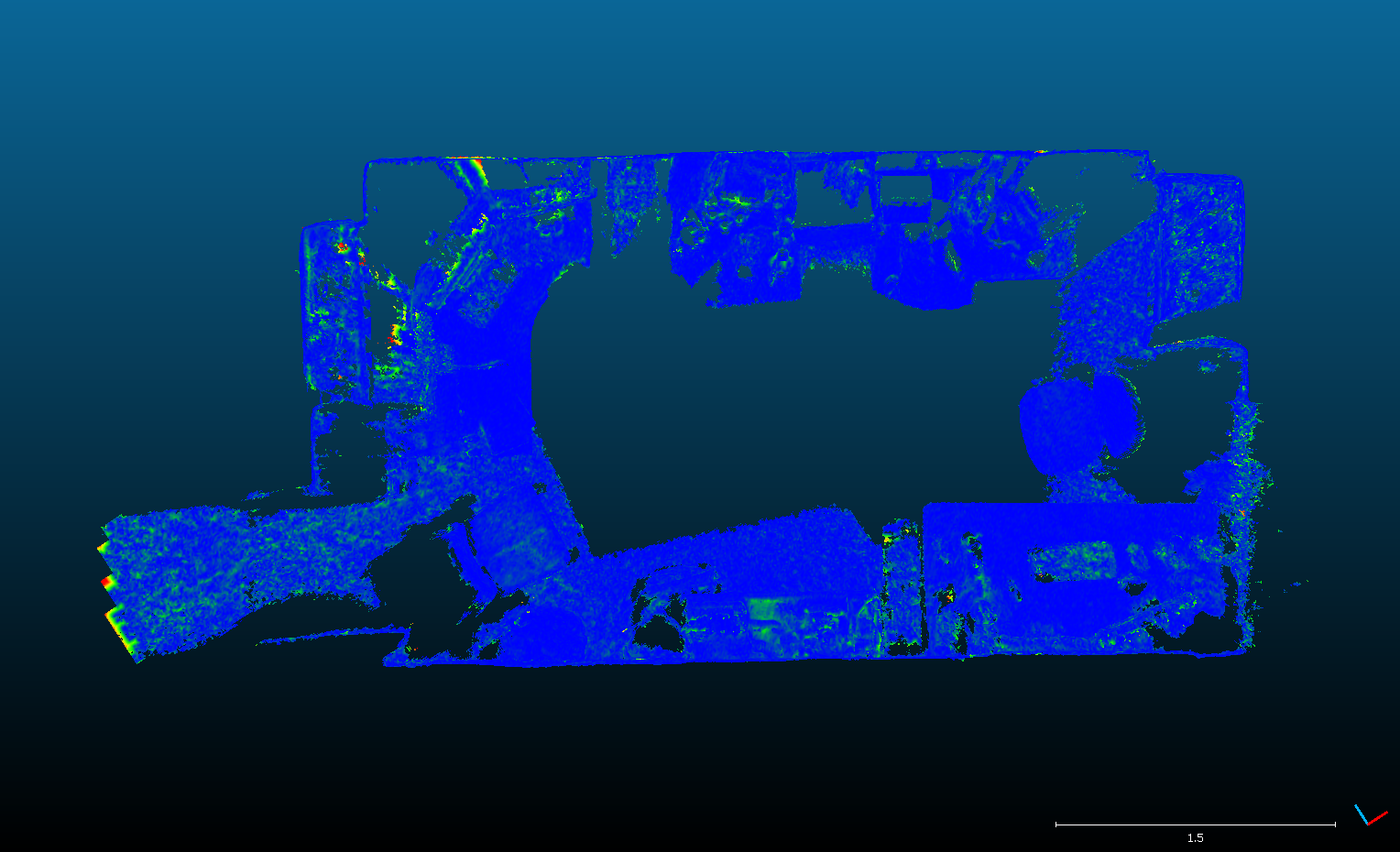}
		\caption{Reference vs.\ Compressed}
	\end{subfigure}%
	\caption{An example showing the impact of frame compression on the reconstruction quality we are able to achieve on the server whilst maintaining real-time frame rates by discarding some frames: (a) a locally-reconstructed reference model of an office; (b) the differences between the reference model and a model reconstructed from the frames that we managed to transmit without using compression; (c) the differences between the reference model and a model reconstructed from the frames that we managed to transmit with compression. The comparisons were made using CloudCompare \cite{CloudCompare}. The errors range from blue = 0cm to red $\ge$ 5cm (or missing). With compression enabled, we are forced to discard far fewer frames, allowing us to achieve a much lower error rate with respect to the reference model.}
	\label{fig:compressionexperiments-wifi}
	\vspace{-1.5\baselineskip}
\end{stusubfig*}

To demonstrate our ability to achieve high-quality collaborative reconstructions, we combined $4$ sequences from the \emph{Flat} subset of our dataset to reconstruct a combined map of a two-bedroom flat (see Figure~\ref{fig:reconstructionquality}). Since a ground-truth reconstruction of the flat was not available (and we did not have access to a LiDAR scanner with which to obtain one), we validated our reconstruction by comparing a variety of measurements made on our combined map with ground-truth measurements made with a laser range finder (a Bosch Professional GLM 40) in the real world. To achieve this, we first converted the voxel-based maps of the sub-scenes produced by InfiniTAM \cite{Prisacariu2014,Prisacariu2017} into mesh-based maps using Marching Cubes \cite{Lorensen1987}. We then applied the relative transforms we had estimated between the sub-scenes during the relocalisation process to these meshes to transform them into a common coordinate system. Finally, we imported the transformed meshes into MeshLab \cite{Cignoni2008}, and used its point-to-point measurement tool to make our measurements. As shown in Figure~\ref{fig:reconstructionquality}, we found that the measurements on our reconstructed model were consistently within $5$cm of the ground truth, indicating that we are able to achieve reconstructions that correspond well to real-world geometry. Further examples, this time showing collaborative reconstructions of multi-storey houses and a three-storey research lab, can be found in \S\ref{sec:dataset}.

\subsection{Effectiveness of Individual Components}
\label{subsec:effectivenessexperiments}

\subsubsection{Frame Compression}
\label{subsubsec:effectiveness-framecompression}

To demonstrate the impact that compressing the RGB-D frames we transmit over the network has on the final reconstruction quality on the server, we used CloudCompare \cite{CloudCompare} to compare a locally-reconstructed reference model of an office with two reconstructions performed from frame data sent over a WiFi connection, one with frame compression disabled and the other with it enabled (see Figure~\ref{fig:compressionexperiments-wifi}). We measured the bandwidth of this WiFi connection to be roughly $8$ Mbits/s, which is less than the roughly $25$ Mbits/s that we needed to transmit the same sequence in compressed form without loss over a wired connection, and much less than the roughly $250$ Mbits/s that would have been needed to transmit the sequence without loss uncompressed. In other words, with compression enabled, we are able to transmit around $1$ in $3$ of the frames in the sequence over WiFi whilst maintaining real-time rates; without compression, this drops to more than $1$ in $30$. As Figure~\ref{fig:compressionexperiments-wifi} shows, this has a significant effect on the resulting reconstruction quality: with compression disabled (b), we lose some parts of the map completely and have a higher error rate across the map as a whole; with compression enabled (c), we manage to reconstruct more or less the entire map, and the error rate is greatly reduced.

\subsubsection{Relocalisation from Synthetic Images}
\label{subsubsec:effectiveness-synthetic}

\begin{table}[!t]
	\centering
	\scriptsize
	\begin{tabular}{ccc}
		\toprule
		\textbf{Sequence} & \textbf{Real Images} & \textbf{Synthetic Images} \\
		\midrule
		Chess & 99.85\% & \textbf{100.00\%} \\
		Fire & \textbf{99.20\%} & 98.85\% \\
		Heads & \textbf{100.00\%} & \textbf{100.00\%} \\
		Office & 99.80\% & \textbf{99.95\%} \\
		Pumpkin & 90.10\% & \textbf{93.00\%} \\
		RedKitchen & 91.34\% & \textbf{99.98\%} \\
		Stairs & 78.30\% & \textbf{88.10\%} \\
		\bottomrule
	\end{tabular}
	\caption{Comparing relocalisation results obtained by rendering synthetic images of the scenes in the 7-Scenes dataset \cite{Shotton2013} from the test poses to those obtained using the real test images, by adapting a regression forest pre-trained on \emph{Office} \cite{Cavallari2017}. Percentages denote proportions of test frames with $\le 5$cm translation error and $\le 5^\circ$ angular error.}
	\label{tbl:syntheticrelocalisations}
	\vspace{-.5\baselineskip}
\end{table}

As described in \S\ref{subsubsec:interagentrelocalisation}, we relocalise the scenes of different agents against each other using synthetic images, rather than the real \mbox{RGB-D} frames originally captured by the agents. This avoids the prohibitive memory cost of storing all RGB-D frames acquired by each agent in RAM (easily hundreds of MBs per agent). However, we might expect using synthetic images to lower relocalisation performance, since we train the relocalisers for each scene using the real input frames.

To verify that this is not a problem, we used the approach of Cavallari et al.\ \cite{Cavallari2017} on the standard 7-Scenes dataset \cite{Shotton2013} to compare the results we are able to obtain with synthetic images to those we can obtain using real images.
Unlike \cite{Cavallari2017}, which used at most $10$ modes in each leaf of the regression forests, we used at most $50$ modes, since we found that this gave better results (in both cases). To test our synthetic approach, we first reconstructed each 7-Scenes sequence from the real training images as normal, then rendered synthetic frames from the testing poses, rather than using the testing images in the dataset. As Table~\ref{tbl:syntheticrelocalisations} shows, our results using synthetic images were at least as good as, and in most cases actually higher than, the results using the real images, verifying that using synthetic images does not decrease relocalisation performance in practice. This is likely because by rendering synthetic images from the reconstructed scene, we implicitly remove noise from the frames to be relocalised, improving pose estimation accuracy.

\subsubsection{Depth Difference Verification}
\label{subsubsec:effectiveness-verification}

\begin{table*}[!t]
	\centering
	\scriptsize
	\begin{tabular}{ccccccccccc}
		\toprule
		\textbf{Scene 1} & \textbf{Scene 2} & \textbf{Total Frames} & \textbf{Relocalised Frames} & \multicolumn{7}{c}{\textbf{Verifier Performance}} \\
		&&&& \textbf{TP} & \textbf{FP} & \textbf{TN} & \textbf{FN} & \textbf{Precision} & \textbf{Recall} & \textbf{Specificity} \\
		\midrule
		f/bathroom2study & f/kitchen2study & 4901 & 2190 & 517 & 931 & 741 & 1 & 35.7\% & 99.8\% & 44.3\% \\
		f/bathroom2study & f/study2sittingroom & 4989 & 519 & 24 & 59 & 436 & 0 & 28.9\% & 100.0\% & 88.1\% \\
		f/kitchen2study & f/study2sittingroom & 5544 & 1254 & 325 & 317 & 609 & 3 & 50.6\% & 99.1\% & 65.8\% \\
		f/study2sittingroom & f/turret2sittingroom & 5895 & 1350 & 555 & 211 & 579 & 5 & 72.5\% & 99.1\% & 73.3\% \\
		\midrule
		h/frontroom2study & h/hall2frontbedroom & 7714 & 1229 & 245 & 90 & 894 & 0 & 73.1\% & 100.0\% & 90.9\% \\
		h/frontroom2study & h/hall2oldkitchen & 8171 & 1307 & 373 & 175 & 757 & 2 & 68.1\% & 99.5\% & 81.2\% \\
		h/hall2frontbedroom & h/hall2oldkitchen & 7141 & 653 & 95 & 106 & 451 & 1 & 47.3\% & 99.0\% & 81.0\% \\
		h/hall2frontbedroom & h/mainbedroom2studio & 6942 & 839 & 90 & 131 & 615 & 3 & 40.7\% & 96.8\% & 82.4\% \\
		\midrule
		\multicolumn{2}{c}{\emph{Average (all scene pairs)}} & -- & -- & -- & -- & -- & -- & 52.1\% & 99.2\% & 75.9\% \\
		\bottomrule
	\end{tabular}
	\caption{Evaluating the effectiveness of the depth difference verification we perform on proposed relocalisations between pairs of sub-scenes (see \S\ref{subsubsec:effectiveness-verification}). We attempt to relocalise every frame of each sub-scene against the other using \cite{Cavallari2017}, and record the number of frames that we were able to relocalise, together with statistics on how many correct/incorrect relocalisations were verified/rejected by the verifier.}
	\label{tbl:verifierexperiments}
\end{table*}

\begin{table*}[!t]
	\centering
	\scriptsize
	\begin{tabular}{ccccccc}
		\toprule
		\textbf{Scene 1} & \textbf{Scene 2} & \textbf{Total Frames} & \textbf{Samples Added} & \textbf{Correct Cluster} & \textbf{Largest Incorrect Cluster} & \textbf{Safety Margin} \\
		\midrule
		f/bathroom2study & f/kitchen2study & 4901 & 1472 & 1317 (89.5\%) & 18 (1.2\%) & 1299 \\
		f/bathroom2study & f/study2sittingroom & 4989 & 82 & 21 (25.6\%) & 18 (22.0\%) & 3 \\
		f/kitchen2study & f/study2sittingroom & 5544 & 589 & 506 (85.9\%) & 5 (0.8\%) & 501 \\
		f/study2sittingroom & f/turret2sittingroom & 5895 & 765 & 759 (99.2\%) & 2 (0.3\%) & 757 \\
		\midrule
		h/frontroom2study & h/hall2frontbedroom & 7714 & 360 & 249 (69.2\%) & 16 (4.4\%) & 233 \\
		h/frontroom2study & h/hall2oldkitchen & 8171 & 592 & 442 (74.7\%) & 10 (1.7\%) & 432 \\
		h/hall2frontbedroom & h/hall2oldkitchen & 7141 & 242 & 192 (79.3\%) & 12 (5.0\%) & 180 \\
		h/hall2frontbedroom & h/mainbedroom2studio & 6942 & 260 & 102 (39.2\%) & 69 (26.5\%) & 33 \\
		\bottomrule
	\end{tabular}
	\caption{Evaluating the extent to which our approach of incrementally clustering the relative transform samples between different pairs of sub-scenes is able to remove outliers and find consistent relocalisations between the sub-scenes. We attempt to relocalise every frame of each sub-scene against the other using \cite{Cavallari2017}, and record the total number of samples that were added (equal to the number of relocalised frames that passed depth difference verification), and the sizes of the correct cluster and the largest incorrect cluster produced by our method in each case (together with the percentages of the samples added these sizes represent). The safety margin for each scene pair refers to the number of consistent erroneous samples that would need to have been added to the largest incorrect cluster to cause it to have been chosen instead of the correct cluster.}
	\label{tbl:clusteringexperiments}
	\vspace{-\baselineskip}
\end{table*}

As mentioned in \S\ref{subsubsec:interagentrelocalisation}, we verify each proposed inter-agent relocalisation of a sub-scene $b$ against a sub-scene $a$ by rendering a synthetic depth image of scene $a$ from the proposed pose and comparing it to the synthetic depth image from scene $b$ that was passed to the relocaliser, masking out pixels that do not have a valid depth in both images. To evaluate the effectiveness of this approach, we first took $8$ pairs of sequences $\{a,b\}$ from our dataset that we were able to successfully relocalise with respect to each other during the normal operation of our approach, and recorded the relative transform between them as the `ground truth' transform for later use. We then attempted to relocalise every frame from scene $b$ using the relocaliser of scene $a$, and vice-versa. Next, for each frame for which the relocaliser proposed a relative transform, we ran our verification step on the proposed transform, thereby classifying it as either \emph{Verified} or \emph{Rejected}. We then compared the proposed transform to the `ground truth' transform, classifying it as \emph{Correct} if it was within $5^\circ$ and $5$cm of the ground truth, and as \emph{Incorrect} otherwise. Finally, we counted Verified/Correct transforms as true positives (TP), Verified/Incorrect transforms as false positives (FP), and similarly for the true negatives (TN) and false negatives (FN), as shown in Table~\ref{tbl:verifierexperiments}.

Our results show that our verifier has an extremely high average recall rate (99.2\%), meaning that it largely manages to avoid rejecting correct transforms. It also has a reasonably good average specificity (75.9\%), meaning that it is fairly good at pruning the number of incorrect transforms we need to deal with. However, a fairly large number of incorrect transforms still manage to pass the verification stage: as mentioned in \S\ref{subsubsec:interagentposeoptimisation}, these are dealt with later by clustering the transforms and only making use of the transforms from the largest cluster. The additional effects of this clustering step are evaluated in \S\ref{subsubsec:effectiveness-clustering}.

\subsubsection{Relative Transform Clustering}
\label{subsubsec:effectiveness-clustering}

As described in \S\ref{subsubsec:interagentposeoptimisation}, we incrementally cluster the relative transform samples $\{{}_a\tilde{T}_b^\mathbf{f}\}$ we add for each pair of sub-scenes $(a,b)$ prior to performing pose graph optimisation to suppress the effect of outliers on the final result. This is achieved by checking each new relative transform sample to see if there is an existing cluster to which it can be added (we specify that this is possible if and only if it is within $20^\circ$ and $10$cm of any existing relative transform in the cluster). If so, we add the sample to the first such cluster we find; if not, we create a new cluster for it.

To evaluate the effectiveness of this approach, we took the same $8$ pairs of sequences $\{a,b\}$ we used to evaluate our depth difference verifier in \S\ref{subsubsec:effectiveness-verification}, and again relocalised every frame from scene $b$ using the relocaliser of scene $a$, and vice-versa. We then counted the number of relative transform samples that were added during this process, and examined the clusters into which they had been collected.
In particular, we compared the size of the largest cluster in each case (i.e.\ the largest `correct' cluster) with the size of the largest cluster whose blended transform (obtained by blending all of the relative transforms in the cluster using dual quaternion blending \cite{Kavan2006}) was not within $20^\circ$ and $10$cm of the blended transform of the correct cluster. We refer to this latter cluster as the largest `incorrect' cluster.
The difference between these two sizes gave us a measure of the \emph{safety margin} of our approach in each case, i.e.\ the number of consistent erroneous samples that would need to have been added to the largest incorrect cluster to cause it to have been chosen instead of the correct cluster.

As our results in Table~\ref{tbl:clusteringexperiments} show, for most pairs of scenes the size of the correct cluster was significantly larger than the size of the largest incorrect cluster, indicating that in practice we are very likely to accumulate $\ge$~$N$ samples from the correct cluster and become confident about it long before we accumulate $\ge$~$N$ samples from an incorrect cluster. For two pairs of scenes (\emph{f/bathroom2study} \& \emph{f/study2sittingroom}, and \emph{h/hall2frontbedroom} \& \emph{h/mainbedroom2studio}), the safety margins were much lower than in the other cases. However, in both cases, the pairs of scenes in question have comparatively low overlap (see the figures showing the green and yellow sequences for \emph{Flat} and \emph{House} in \S\ref{sec:dataset}). Moreover, whilst the blended transforms of the correct cluster and the largest incorrect cluster in each case were not within $20^\circ$ and $10$cm of each other, a manual inspection of the relevant transforms showed that they were still comparatively close, meaning that the safety margin before hitting a cluster with a grossly incorrect transform is in practice somewhat higher in each case.

\subsection{Scalability}
\label{subsec:scalabilityexperiments}

\subsubsection{Time}
\label{subsubsec:scalability-interactive}

\begin{figure}[!t]
	\begin{center}
		\includegraphics[width=\linewidth]{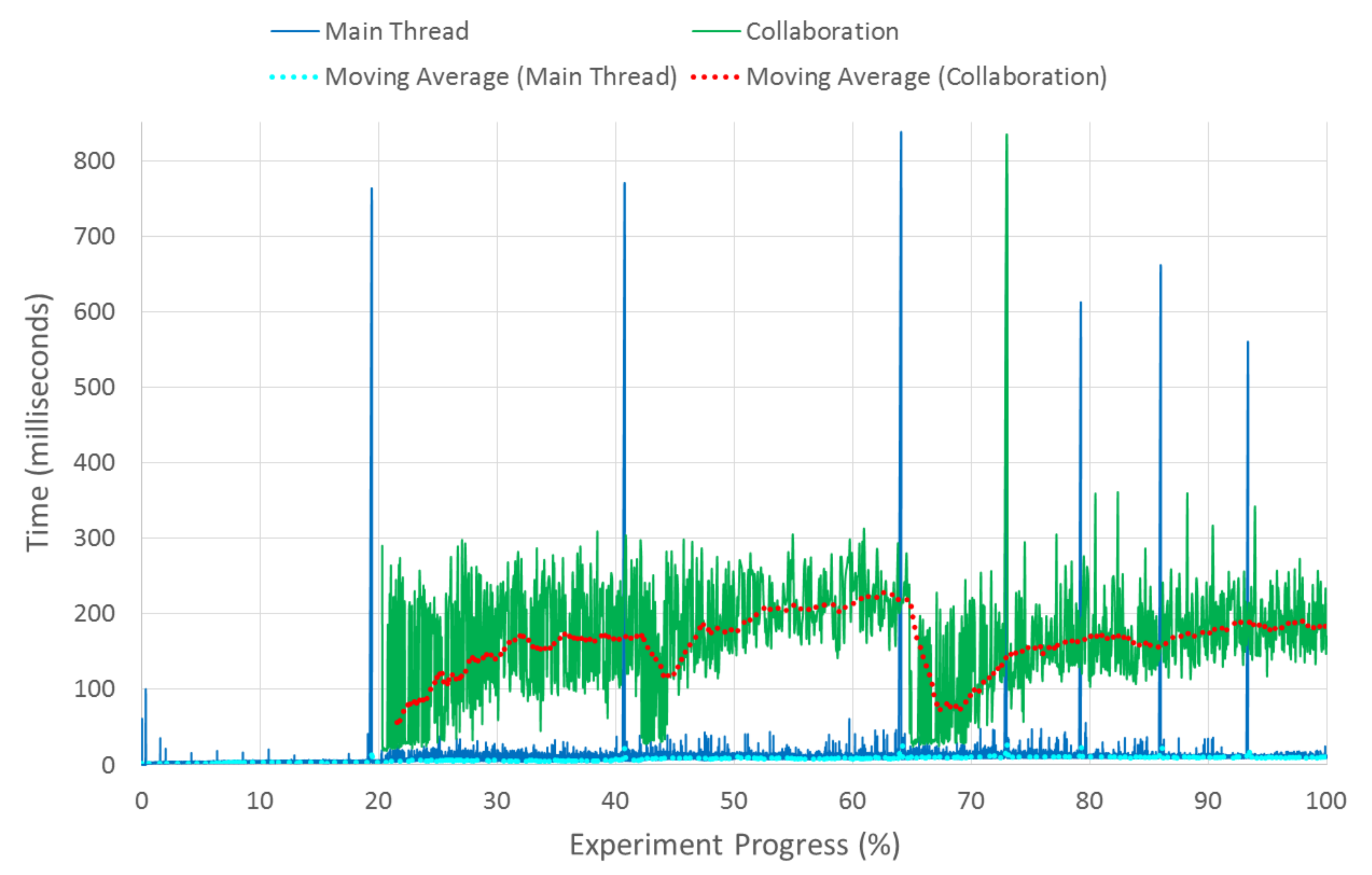}
		\caption{How the time taken by our approach (i) for sub-scene fusion / overall scene rendering (on the main thread), and (ii) for attempting background relocalisations between different sub-scenes (on a separate thread), changed whilst collaboratively reconstructing an office using a server and $4$ Android clients, connected via a WiFi router (see \S\ref{subsubsec:scalability-interactive}).}
		\label{fig:interactivity}
	\end{center}
	\vspace{-1.5\baselineskip}
\end{figure}

To evaluate how well the time taken by our approach scales when multiple clients are collaborating to reconstruct a scene, we timed an interactive collaborative reconstruction of an office using a server and four Android clients, connected via a WiFi router (the server was directly connected to the router via Ethernet for improved bandwidth, whilst the clients were connected wirelessly). In particular, we timed how long it took to process each frame (i.e.\ fusing a sub-scene for each client and rendering the overall scene) on the main thread, and how long it took to perform relocalisations between the different sub-scenes on a separate thread.

The results of this process are shown in Figure~\ref{fig:interactivity}. The four clients connected one at a time at suitably spaced intervals, before later disconnecting again (see the corresponding time spikes, caused by allocating / deallocating memory for the different clients). As Figure~\ref{fig:interactivity} shows, the cost of fusing the sub-scenes and rendering the scene was consistently low (generally less than $50$ms) throughout the experiment, allowing the user to continue to view the collaborative reconstruction interactively from different angles in real time on the server even with multiple clients connected.

Relocalisations between the sub-scenes were started once the second client connected (roughly 19\% of the way through the experiment). As expected, since our approach schedules a single relocalisation at a time, the cost of attempting relocalisations between the sub-scenes stayed bounded over time (generally under $300$ms, including the time taken to render the synthetic images and perform depth difference verification). Interestingly, there was a drop in the time taken per relocalisation after each client connected, caused by the online relocaliser for the new client taking some time to train (initially, each new relocaliser has empty leaves, making relocalisation attempts faster but less likely to be successful; the time taken and the likelihood of a successful relocalisation increase as the relocaliser is trained). The spike in the relocalisation time roughly 73\% of the way through the experiment was caused by the synthetic image rendering for the active relocaliser (which happens on the main GPU, where the sub-scenes are stored) blocking whilst the memory for the first client to disconnect was deallocated.

Overall, these results show that, from a time perspective, our approach should be able to scale to a much larger number of clients without impeding interactivity (we were only able to test with the four Android clients we had available, but given that the time taken by our approach increases by only a small amount for each additional client added, it should in practice remain reasonable even when handling a much larger number of clients). This is to be expected, given the efficiency of modern voxel fusion frameworks such as InfiniTAM \cite{Prisacariu2017}, and the way in which we schedule our relocalisation attempts in the background to avoid blocking the primary GPU. In practice, the main limitation of our approach is currently GPU memory usage, as we show in the next section.

\subsubsection{Memory Usage}
\label{subsubsec:scalability-batch}

\stufig{width=\linewidth}{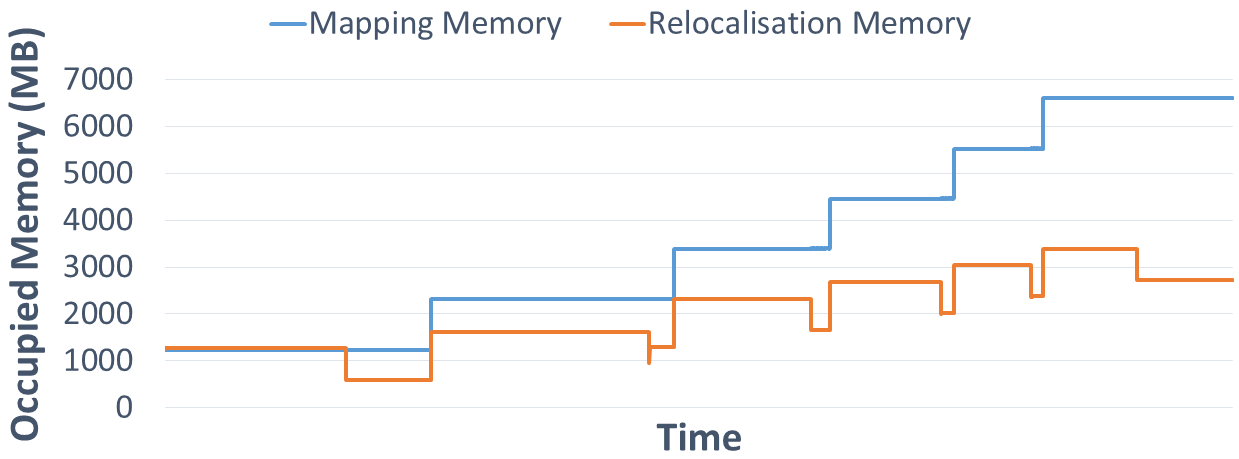}{The GPU memory we use during a collaborative reconstruction of the $6$ \emph{Priory} sequences from our dataset.}{fig:scalability}{!t}

To evaluate the scalability of our approach's GPU memory usage, we performed an offline collaborative reconstruction of the $6$ \emph{Priory} sequences from our dataset (these represent a three-story house: see \S\ref{sec:dataset}) and examined the memory usage on the server as the number of clients increased. To maximise the number of agents we were able to handle, we added new sub-scenes to the server one at a time and deleted the training data used by each sub-scene's relocaliser once it had been fully trained: this limited the maximum memory used by the relocalisers, leaving us bounded primarily by the size of the reconstructed voxel scenes. Figure~\ref{fig:scalability} shows that the final GPU memory we use for each forest-based relocaliser \cite{Cavallari2017} is $< 500$MB, meaning that we can potentially handle around $24$ relocalisers on a $12$GB GPU such as an NVIDIA Titan X (the actual number may be slightly less than this due to driver overhead).
The memory used by the voxel scenes is currently the bottleneck preventing the scaling of our approach to more agents: currently, each scene takes up just over $1$GB of memory, limiting us to around $11$ agents on a Titan X (assuming the relocalisers are stored on a secondary GPU). To scale further, we could reduce the memory used by meshing each scene with Marching Cubes \cite{Lorensen1987} and discarding the voxel maps.

\subsection{Overall Timings for Collaborative Reconstruction}
\label{subsec:timingexperiments}

\begin{table}[!t]
	\centering
	\scriptsize
	\begin{tabular}{ccccc}
		\toprule
		& \textbf{Flat} & \textbf{Priory} & \textbf{House} & \textbf{Lab} \\
		\midrule
		\textbf{\# Sequences} & 4 & 6 & 4 & 4 \\
		\textbf{Longest Sequence: \# Frames} & 3079 & 1640 & 4372 & 3188 \\
		\textbf{Longest Sequence: Time (s)} & 615.8 & 328.0 & 874.4 & 637.6 \\
		\textbf{Average Collaboration Time (s)} & 15.8 & 38.4 & 107.4 & 272.0 \\
		\midrule
		\textbf{Average Total Time (s)} & 631.6 & 366.4 & 981.8 & 909.6 \\
		\bottomrule
	\end{tabular}
	\caption{The total times taken to collaboratively reconstruct four different subsets of our dataset (see \S\ref{subsec:timingexperiments}).}
	\label{tbl:timings}
	\vspace{-\baselineskip}
\end{table}

To evaluate how long it takes to produce consistent reconstructions using our approach, we computed the average times taken to collaboratively reconstruct four different subsets of our dataset (see Table~\ref{tbl:timings}). We computed the time taken to capture the sequences in each subset as $\frac{1}{5}$ times the length of the longest sequence, assuming parallel capturing at $5$Hz. The average collaboration time is the time taken to relocalise the agents against each other and compute optimised global poses for their maps. To account for the random selection of frames to relocalise, we globally mapped each subset $5$ times and reported the average time in the table. The average total time is from the start of the capturing process to the output of a globally consistent map: this was under half an hour for all subsets we tested.

\subsection{Adaptability to Different Clients}
\label{subsec:adaptabilityexperiments}

To demonstrate our approach's ability to work with clients implemented in different ways, we used BundleFusion \cite{Dai2017}, a state-of-the-art real-time globally consistent reconstruction approach that can close loops and re-integrate the scene surface on the fly, to recapture the two-bedroom flat that forms the \emph{Flat} subset of our dataset. Since BundleFusion can only produce accurate camera poses once loops have actually been closed, we reconstructed each of the four sub-scenes involved ahead of time and then combined them using the batch mode of our system. The results of this process are shown in Figure~\ref{fig:bundlefusion}. In comparison to reconstruction using the Android client, we found that reconstruction using BundleFusion required far more care to be taken during the capture process, since its visual-only tracker is far more prone to fail than one based on visual-inertial odometry. (If an IMU is available, BundleFusion's tracker can in principle be replaced with one based on visual-inertial odometry, e.g.\ \cite{Bloesch2015}, but doing so is non-trivial and beyond the scope of the present paper.) Nevertheless, at \emph{Flat} scale, we found BundleFusion to be a viable replacement for our Android client.

\section{Conclusion}

\stufig{width=\linewidth}{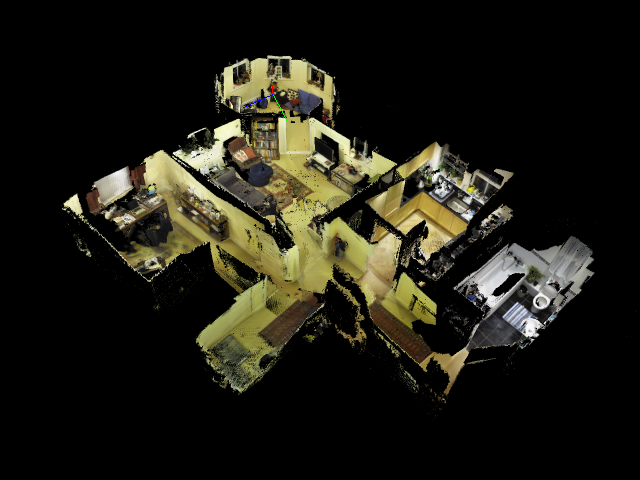}{A collaborative reconstruction of a two-bedroom flat, produced by running our approach in batch mode on four sub-scenes captured using BundleFusion \cite{Dai2017}.}{fig:bundlefusion}{!t}

In this paper, we have shown how to enable multiple agents to collaborate interactively to reconstruct dense, volumetric models of 3D scenes. Existing collaborative mapping approaches have traditionally suffered from an inability to trust the local poses produced by their mobile agents, forcing them to perform costly global optimisations (e.g.\ on a server) to ensure a consistent map, and limiting their ability to perform collaborative dense mapping interactively.

By leveraging recent advances to construct rigid local sub-scenes that do not need further refinement, and joining them using a state-of-the-art regression-based relocaliser, we avoid expensive global optimisations, opting only to refine the relative poses between individual agents' maps. Our system allows multiple users to collaboratively reconstruct consistent dense models of entire buildings in under half an hour using only consumer-grade hardware, providing a low-cost, time-efficient and interactive alternative to existing whole-building reconstruction methods based on panorama scanners, and making it easier than ever before for users to capture detailed 3D scene models at scale.

\newpage

\appendix

\part*{Supplementary Material}

\begin{table}[H]
	\centering
	\footnotesize
	\begin{tabular}{ccc}
		\toprule
		\textbf{Sequence} & \textbf{Frame Count} & \textbf{Capture Time (s)} \\
		\midrule
		f/bathroom2study & 2173 & 434.6 \\
		f/kitchen2study & 2728 & 545.6 \\
		f/study2sittingroom & 2816 & 563.2 \\
		f/turret2sittingroom & 3079 & 615.8 \\
		\midrule
		h/frontroom2study & 4372 & 874.4 \\
		h/hall2frontbedroom & 3342 & 668.4 \\
		h/hall2oldkitchen & 3798 & 759.6 \\
		h/mainbedroom2studio & 3600 & 720.0 \\
		\midrule
		p/bath2office & 1319 & 263.8 \\
		p/bed2office & 1518 & 303.6 \\
		p/dining2guest & 1232 & 246.4 \\
		p/guest2bath & 1315 & 263.0 \\
		p/kitchen2dining & 1142 & 228.4 \\
		p/living2dining & 1640 & 328.0 \\
		\midrule
		l/atrium & 1709 & 341.8 \\
		l/firstfloor & 3188 & 637.6 \\
		l/groundfloor & 1985 & 397.0 \\
		l/secondfloor & 2146 & 429.2 \\
		\bottomrule
	\end{tabular}
	\caption{The sequences in each subset of our dataset.}
	\label{tbl:dataset}
\end{table}

\section{Our Dataset}
\label{sec:dataset}

Our dataset comprises $4$ different subsets -- Flat, House, Priory and Lab -- each containing a number of different sequences that can be successfully relocalised against each other. The name of each sequence is prefixed with a simple identifier indicating the subset to which it belongs: f/ for Flat, h/ for House, p/ for Priory and l/ for Lab. Basic information about the sequences in each subset, their frame counts and capture times can be found in Table~\ref{tbl:dataset}. Illustrations of the sequences and how they fit together to make the combined scenes are shown in Figures~\ref{fig:frenchay-subset} to \ref{fig:lab-subset}.

Each sequence was captured at $5$Hz using an Asus ZenFone AR augmented reality smartphone, which produces depth images at a resolution of $224 \times 172$, and colour images at a resolution of $1920 \times 1080$. To improve the speed at which we were able to load sequences from disk, we resized the colour images down to $480 \times 270$ (i.e.\ 25\% size) to produce the collaborative reconstructions we show in the paper, but we nevertheless provide both the original and resized images as part of the dataset. We also provide the calibration parameters for the depth and colour sensors, the 6D camera pose at each frame, and the optimised global pose produced for each sequence when running our approach on all of the sequences in each subset. Finally, we provide a pre-built mesh of each sequence, pre-transformed by its optimised global pose to allow the sequences from each subset to be loaded into MeshLab \cite{Cignoni2008} or CloudCompare \cite{CloudCompare} with a common coordinate system.

\section{Additional Implementation Details}

In this section, we describe some relevant implementation details for those wanting to reimplement our approach, namely the inner workings of the pooled queue data structure mentioned in \S\ref{subsec:frametransmission}, the way in which we can render a global map of all of the sub-scenes we have reconstructed, and the way in which those sub-scenes can be fused offline to produce a single global reconstruction. These details can be skipped by more casual readers.

\subsection{Pooled Queue Data Structure}
\label{subsec:pooledqueue}

A \emph{pooled queue} is a data structure that pairs a normal queue $Q$ of objects of type $T$ with a pool $P$ of reusable $T$ objects, with the underlying goal of minimising memory reallocations for performance reasons. Normal queues conventionally support the following range of operations: \texttt{empty} checks whether the queue is empty, \texttt{peek} gets the object (if any) at the front of the queue, \texttt{pop} removes the object (if any) at the front of the queue, \texttt{push} adds an object to the back of the queue, and \texttt{size} gets the number of objects on the queue. Pooled queues support the same range of operations, but their implementations of \texttt{push} and \texttt{pop} are necessarily complicated by interactions with the pool.

The \texttt{pop} operation is the more straightforward of the two: this simply removes the object (if any) at the front of the queue and returns it to the pool (with appropriate synchronisation in a multi-threaded context).

The \texttt{push} operation is more complicated, and we divide it into two parts: \texttt{begin\_push} and \texttt{end\_push}. The \texttt{begin\_push} operation first checks whether a reusable object is currently available from the pool. If so, it removes it from the pool and returns a \emph{push handler} encapsulating the object to the caller. The caller then modifies the object, after which the \emph{push handler} calls \texttt{end\_push} to actually push the object onto the queue. If a reusable object is not available from the pool when \texttt{begin\_push} is called, we have a range of options, which are available as policies in our implementation: (i) discard the object we are trying to push onto the queue, (ii) grow the pool by allocating a new object that can then be reused, (iii) remove a random element from the queue and return it to the pool so that it can be reused, or (iv) wait for another thread to pop an object from the queue and return it to the pool. In practice, we found the discard strategy to work best for our approach: growing the pool has the disadvantage that the memory usage can grow without bound over time, waiting makes the client and/or server (as the case may be) less interactive, and removing a random element from the queue functions much like discarding but with different frames.

\begin{stusubfig*}{!t}
	\begin{subfigure}{.48\linewidth}
		\centering
		\includegraphics[width=\linewidth]{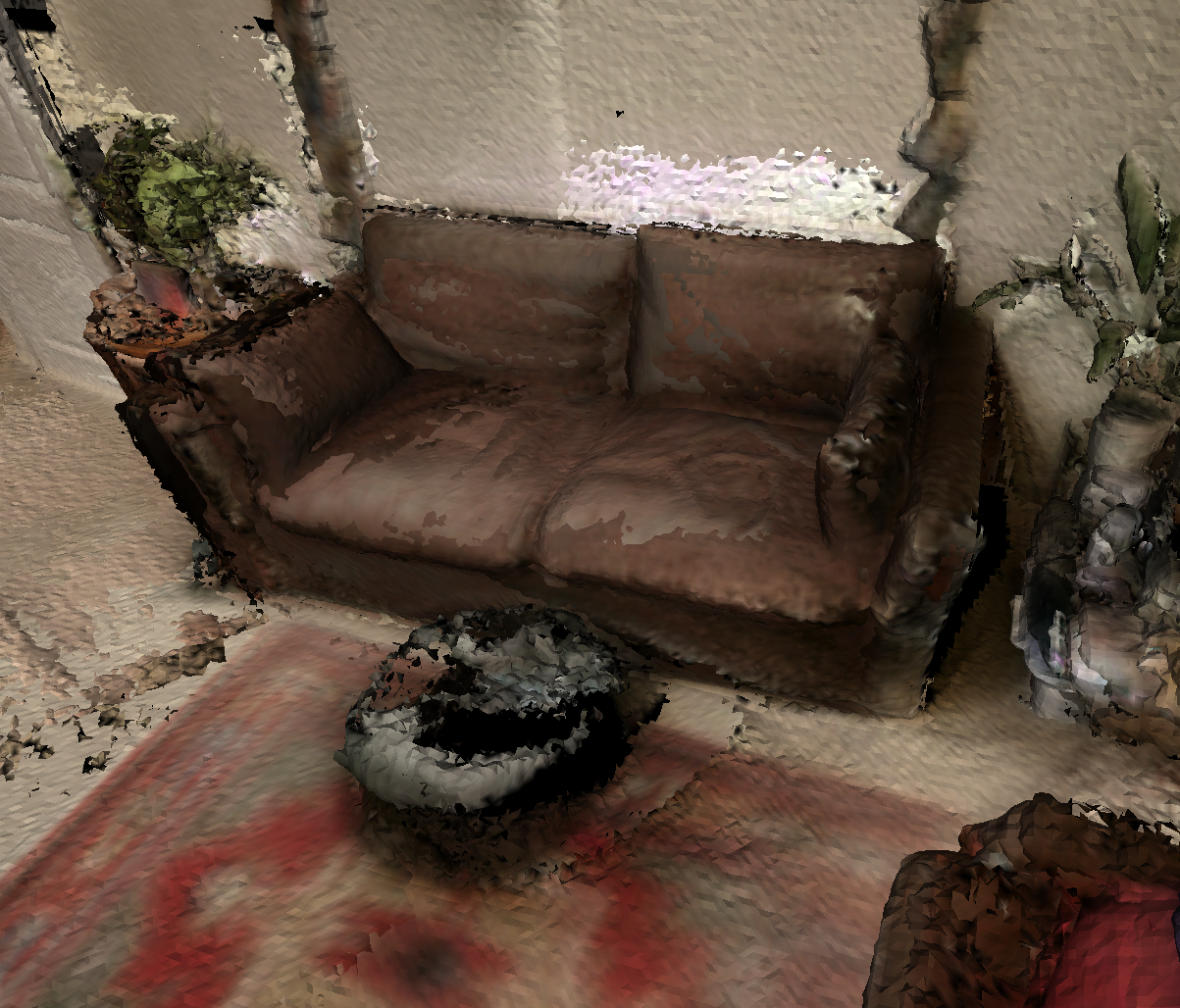}
		\caption{Rendering a global map of the sub-scenes with per-pixel depth testing}
	\end{subfigure}%
	\hspace{4mm}%
	\begin{subfigure}{.48\linewidth}
		\centering
		\includegraphics[width=\linewidth]{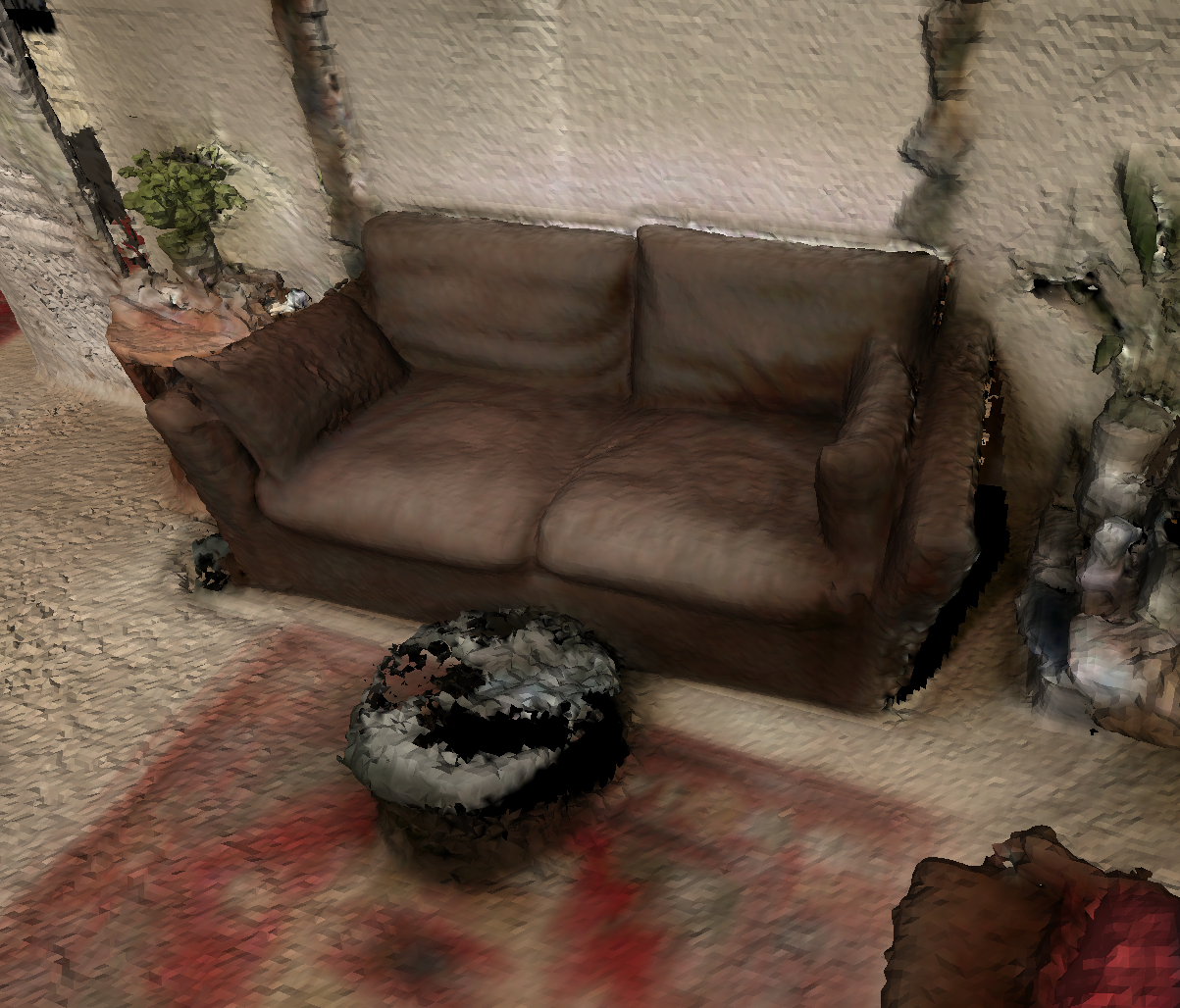}
		\caption{Rendering the single global reconstruction produced by fusing the sub-scenes together}
	\end{subfigure}%
	\caption{Fusing the sub-scenes into a single global reconstruction, as an offline post-processing step, naturally resolves inconsistencies between the individual sub-scenes, e.g.\ those involving the sofa, wall, left-hand table plant and floor in this example (in which we fused all four sub-scenes in the Flat subset of our dataset). Both reconstructions were rendered using MeshLab \cite{Cignoni2008}, so as to achieve a consistent viewpoint.}
	\label{fig:globalreconstruction}
	\vspace{-1.5\baselineskip}
\end{stusubfig*}

\subsection{Global Map Rendering}

To render a global map of all of the sub-scenes we have reconstructed from a global pose $T_g \in \mathit{SE}_3$, we first compute (for each agent $a$) the corresponding local pose $T_a$ in $a$'s local coordinate system:
\begin{equation}
T_a = T_g \; {}_g\hat{T}_a
\end{equation}
Using the standard raycasting approach for a voxel-based scene \cite{Prisacariu2017}, we then render synthetic colour and depth images (at $1280 \times 960$ resolution) for each agent from the local pose we have computed for it, which we call $C_a$ and $D_a$ respectively. Finally, we set the colour of each pixel in the output image $\mathcal{O}$ based on per-pixel depth testing between the agents, i.e.
\begin{equation}
\mathcal{O}(\mathbf{x}) = C_{a^*(\mathbf{x})}(\mathbf{x}), \mbox{ where } a^*(\mathbf{x}) = \argmin_a D_a(\mathbf{x}),
\end{equation}
where $\mathbf{x}$ ranges over the domain of $\mathcal{O}$.

\subsection{Producing a Single Global Reconstruction}
\label{subsec:singleglobalreconstruction}

To preserve the possibility of moving the sub-scenes of individual agents relative to each other during collaborative reconstruction (in particular, as part of the pose graph optimisation process), we do not merge them into a single global reconstruction online. However, our system does provide a mechanism for fusing the sub-scenes into a single reconstruction after the fact.

To achieve this, we save both the sequences used to reconstruct the sub-scenes for the agents and the global poses $\{{}_g\hat{T}_a\}$ determined by the pose graph optimiser to disk at the end of the collaborative reconstruction process. We then read the sequences and global poses back in, and feed the sequences' frames sequentially through a standard TSDF fusion pipeline \cite{Prisacariu2017}, transforming each frame's pose by its sequence's global pose immediately prior to fusion. This process produces a single global model that can then be used as normal for downstream tasks. Any inconsistencies between the individual sub-scenes (in overlapping areas) are naturally resolved by the implicit surface used in the fusion process.

Figure~\ref{fig:globalreconstruction} shows an example of rendering a fused model produced by this process, in comparison to simply rendering a global map of the corresponding sub-scenes (without fusing them) as described in the previous sub-section. The fused model resolves minor inconsistencies where the sub-scenes overlap, but the individual sub-scenes can no longer be moved around once fusion has taken place.

\section{Depth Sensor Noise Characteristics}

As with all approaches to 3D reconstruction from RGB-D cameras, the reconstruction quality our approach is able to achieve is partly a function of the quality of the depth produced by the sensors we use. For that reason, we include here a brief discussion of the noise characteristics of several popular consumer depth sensors (Kinect v1, PrimeSense, and Kinect v2) that can be used with our system. For a more complete discussion, the interested reader may wish to consult the mentioned references. Unfortunately, we are not aware of any existing public study of the noise characteristics of the depth sensor embedded in the Asus ZenFone AR.

The noise characteristics of the original Kinect v1 sensor (which is based on structured light) were modelled by Nguyen et al.\ \cite{Nguyen2012}, who determined the axial depth noise by freely rotating a planar target around a vertical axis in front of a fixed Kinect, fitting a plane to the resulting point cloud, and then examining the differences between the raw and fitted depths. The lateral depth noise was determined by examining vertical edges in the depth map. They found that the axial noise increases quadratically with depth, whereas the lateral noise increases only linearly. Mallick et al.\ \cite{Mallick2014} obtained similar results. The PrimeSense sensors are based on the same structured light approach as the Kinect v1 (which uses PrimeSense's technology \cite{Zanuttigh2016}). The noise characteristics of both devices are thus similar.

Wasenm{\"u}ller et al.\ \cite{Wasenmueller2016ACCV} undertook a detailed study for the Kinect v2 sensor (which is based on time-of-flight principles). They found that in contrast to the Kinect v1, whose accuracy decreases exponentially with increasing distance, the accuracy of the new sensor is constant with depth (roughly a $-18$mm offset from the true depth). However, the depth accuracy of the Kinect v2 is dependent on temperature, with the authors recommending that it be pre-heated in advance to achieve best results. Moreover, unlike the Kinect v1, the accuracy of the Kinect v2's depth is affected by scene colour. Yang et al.\ \cite{Yang2015} also studied the depth accuracy of the Kinect v2, using a cone model to represent its accuracy distribution. In addition to studying the behaviour of an individual Kinect v2, they proposed a method for improving depth accuracy when three separate sensors can be used. Finally, Fankhauser et al.\ \cite{Fankhauser2015} studied the Kinect v2's usefulness in the context of mobile robot navigation.

\stufigstar{width=\linewidth}{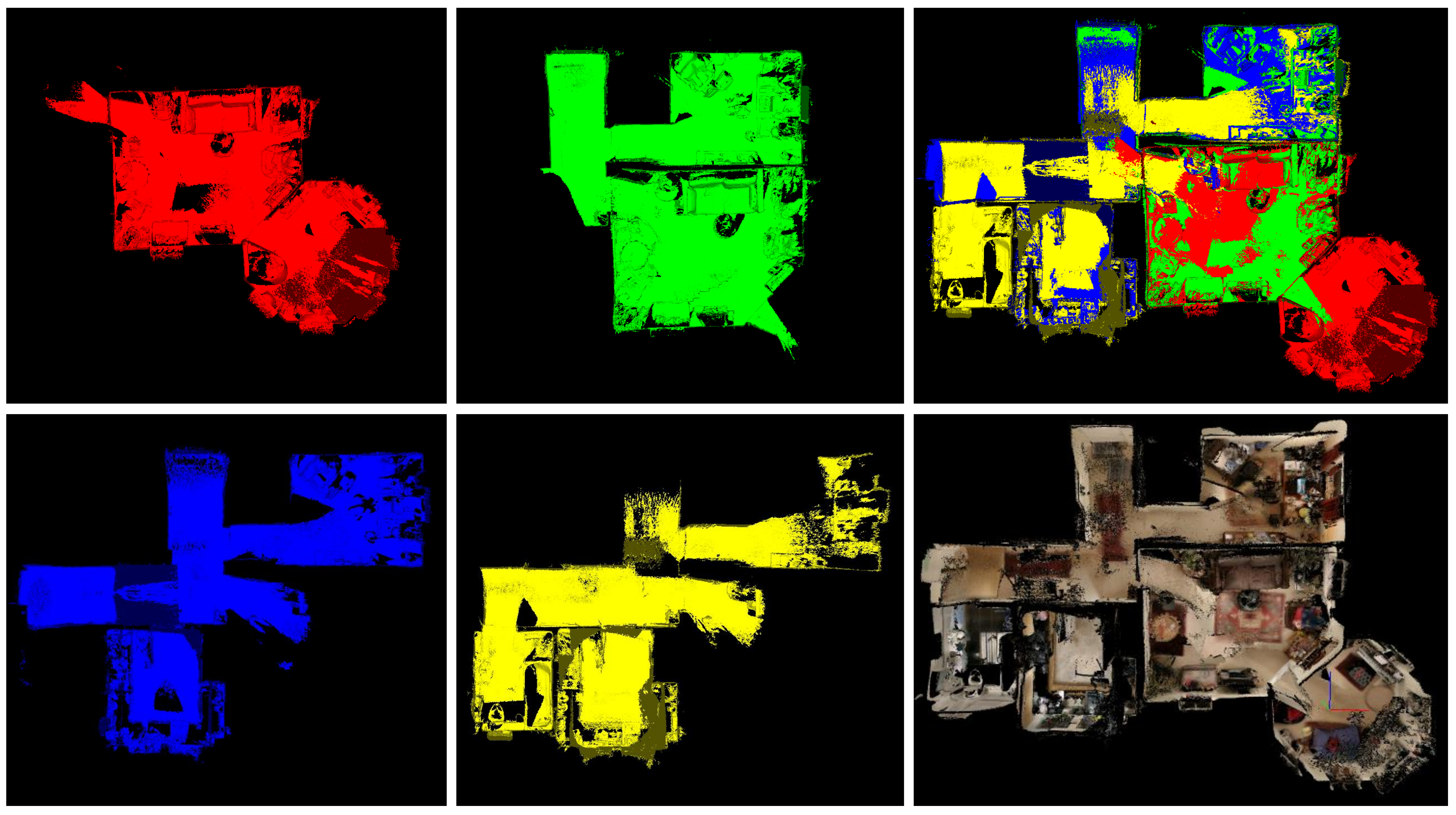}{A collaborative reconstruction of the $4$ Flat sequences in our dataset (\emph{f/turret2sittingroom}, \emph{f/study2sittingroom}, \emph{f/kitchen2study} and \emph{f/bathroom2study}). These collectively represent a single-storey, two-bedroom flat. The mono-colour images show the individual sub-scenes we reconstruct from each sequence; the other images show the combined map.}{fig:frenchay-subset}{!p}

\stufigstar{width=\linewidth}{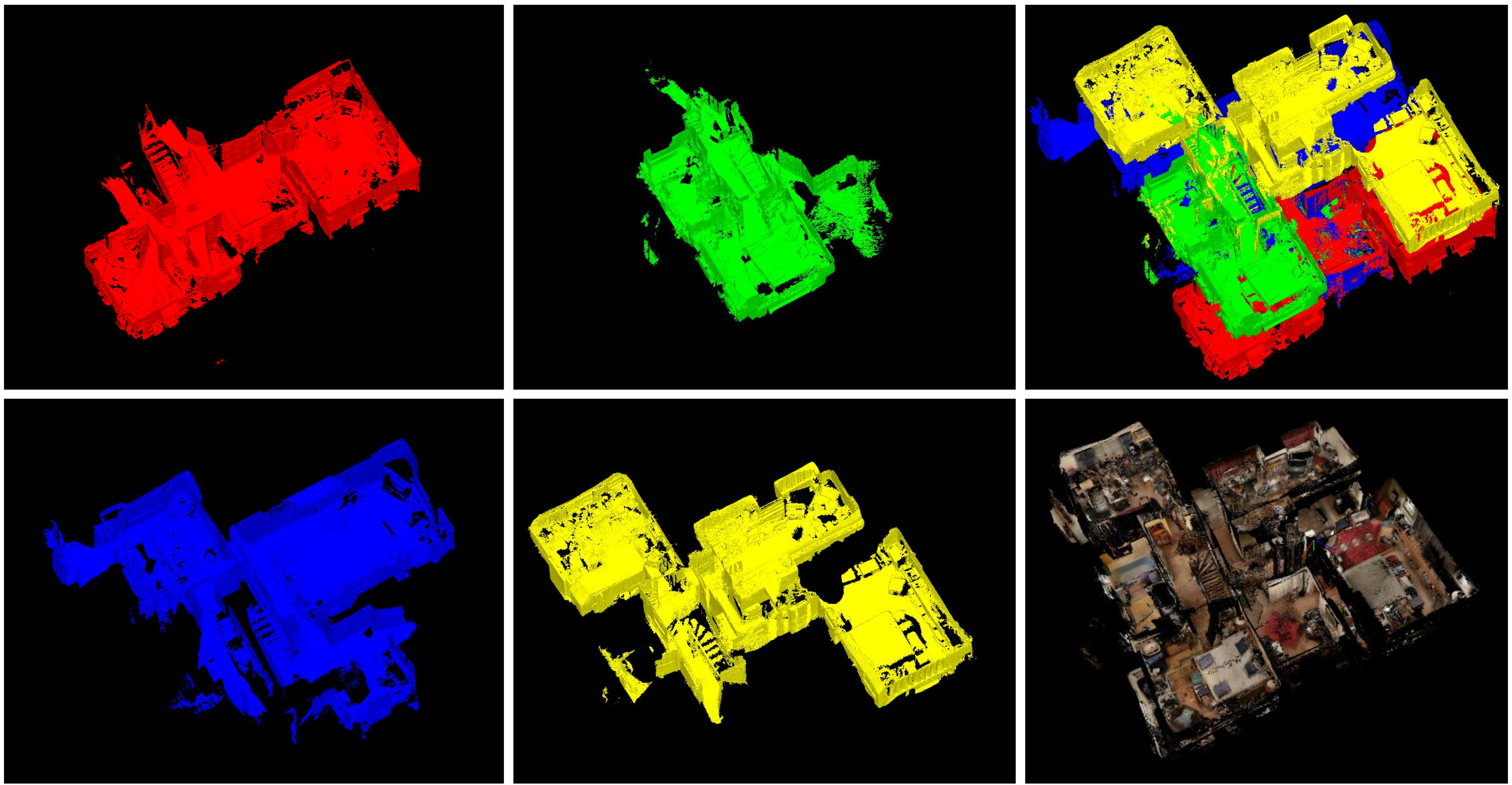}{A collaborative reconstruction of the $4$ House sequences in our dataset (\emph{h/frontroom2study}, \emph{h/hall2frontbedroom}, \emph{h/hall2oldkitchen} and \emph{h/mainbedroom2studio}). These collectively represent a two-storey, four-bedroom house. The mono-colour images show the individual sub-scenes we reconstruct from each sequence; the other images show the \mbox{combined map.}}{fig:dukeswood-subset}{!p}

\stufigstar{width=.925\linewidth}{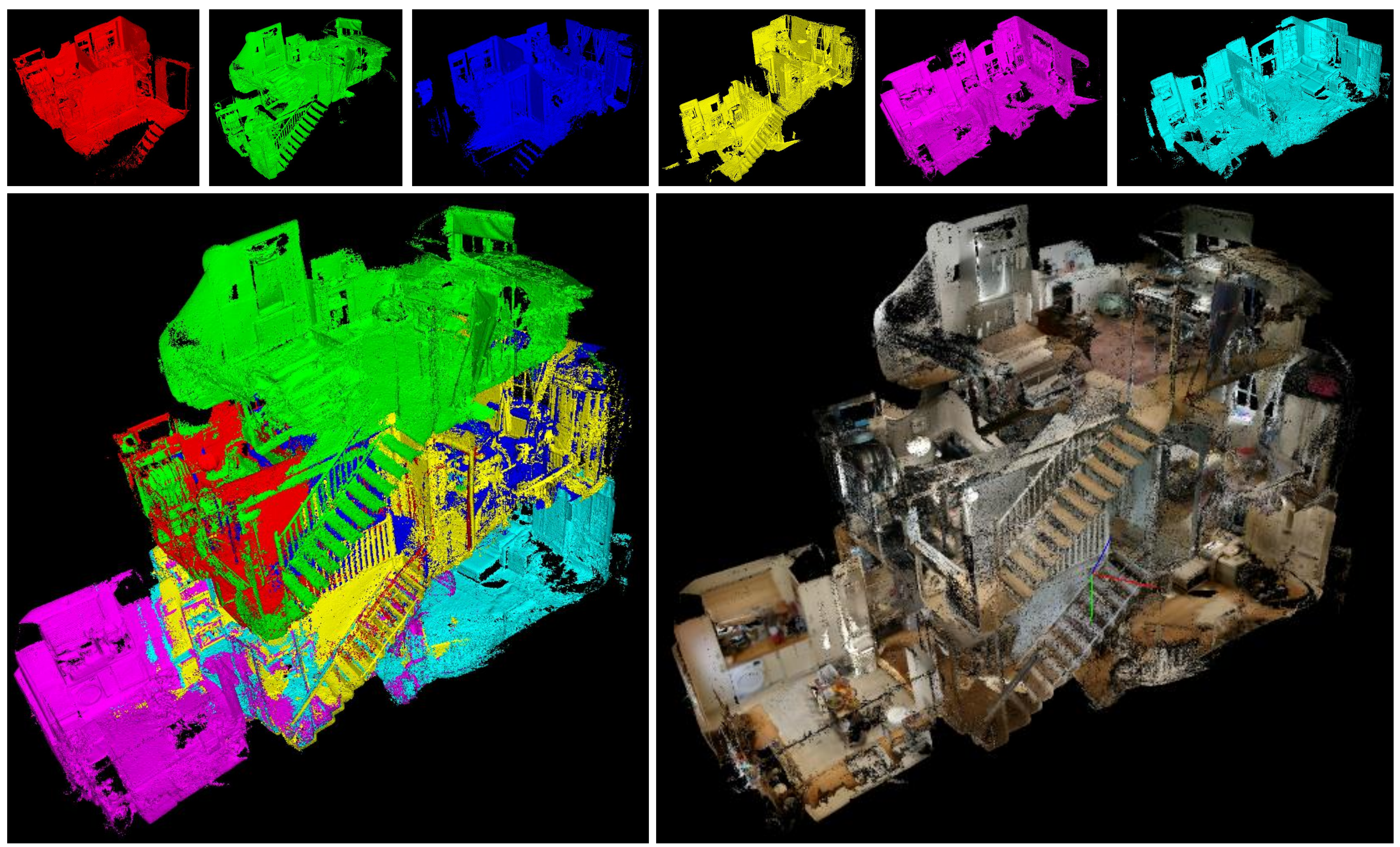}{A collaborative reconstruction of the $6$ Priory sequences in our dataset (\emph{p/bath2office}, \emph{p/bed2office}, \emph{p/guest2bath}, \emph{p/dining2guest}, \emph{p/kitchen2dining} and \emph{p/living2dining}). These collectively represent a three-storey house. The mono-colour images show the individual sub-scenes we reconstruct from each sequence; the other images show the combined map.}{fig:priory-subset}{!p}

\stufigstar{width=.925\linewidth}{lab-subset}{A collaborative reconstruction of the $4$ Lab sequences in our dataset (\emph{l/groundfloor}, \emph{l/firstfloor}, \emph{l/atrium} and \emph{l/secondfloor}). These collectively represent a three-storey research lab. The mono-colour images show the individual sub-scenes we reconstruct from each sequence; the other images show the combined map.}{fig:lab-subset}{!p}

\acknowledgments{This work was supported by Innovate UK/CCAV project 103700 (StreetWise), the EPSRC, ERC grant ERC-2012-AdG 321162-HELIOS, EPSRC grant Seebibyte EP/M013774/1 and EPSRC/MURI grant EP/N019474/1. We would also like to thank Manar Marzouk and Maria Anastasia Tsitsigkou for their help with the collaborative dataset collection, and Oscar Rahnama, Tom Joy, Daniela Massiceti, Nantas Nardelli, Mohammad Najafi and Adnane Boukhayma for their help with the experiments.}

\bibliographystyle{abbrv-doi}

\bibliography{distmappaper}
\end{document}